\documentclass[12pt,letterpaper]{article}
\usepackage{times}
\usepackage{epsfig}
\usepackage{balance}
\usepackage{rotating}
\usepackage{subfig}
\usepackage{graphicx}
\usepackage{tabularx}
\usepackage{color}
\usepackage{array}
\usepackage[margin=0.75in]{geometry}
\usepackage{amsmath,amsthm, amssymb, latexsym, enumerate}
\usepackage{biblatex}

\newtheorem{theorem}{Theorem}

\theoremstyle{remark}
\newtheorem*{remark}{Remark}

\theoremstyle{definition}
\newtheorem{definition}{Definition}[section]
\newtheorem{corollary}{Corollary}[section]

\usepackage{mathtools}

\begin{document}

\title{From Boundaries to Bumps:\\
when closed (extremal) contours are critical}
\author{Benjamin Kunsberg \\
Applied Mathematics \\ Brown University \\
Providence, RI \and
Steven W. Zucker\thanks{Research supported by the Paul G. Allen Frontiers Group, the Simons Collaboration on the Global Brain, and NSF CRCNS AWD0001607.}\\
Computer Science, Biomedical Engineering \\ Yale University \\
New Haven, CT \\
steven.zucker@yale.edu}

\maketitle

\abstract{Invariants underlying shape inference are elusive: a variety of shapes can give rise to the same image, and a variety of images can be rendered from the same shape. The occluding contour is a rare exception: it has both image salience, in terms of isophotes, and surface meaning, in terms of surface normal.  We relax the notion of occluding contour to define closed extremal curves, a new shape invariant that exists at the topological level. They surround bumps, a common but ill-specified interior shape component, and formalize the qualitative nature of bump perception. Extremal curves are biologically computable, unify shape inferences from shading, texture, and specular materials, and predict new phenomena in bump perception.}

\section{Introduction}

When describing a shape, it is natural to use terms such as bumps, dents, valleys and ridges. But what, precisely, is a bump? The Oxford English Dictionary defines it as
a protuberance on a level surface, as in {\em a bump in the road}. While intuitively pleasing, this vagueness hides basic questions in shape perception; we pose them in terms of bumps. 

\begin{itemize}
    \item[1] {\em Bumps are global objects}; they are defined not at a point, such as the curvature at the peak, but over a neighborhood. Like a mountain, a bump is a collection of material that builds to a peak; they can be climbed from many sides.  How can this global aspect of a bump be characterized? We shall exploit the observation that the level sets, in climbing a bump, are nested and increasing, and shall use it to identify a novel relationship to flow pattern perception.
    \item[2] {\em Bumps are qualitative}. While bumps have a boundary, precisely where it lies is less clear. Evidence suggests that shape perception is qualitative: that is, while different subjects agree on certain basic properties of shape, they disagree on quantitative details. We shall encompass this qualitative aspect of shape inferences in a topological representation.
    \item[3] {\em Bumps exist in the interior} of a shape. The occluding boundary delimits the full extent of an object. We shall show that bumps have a description that is a relaxation of the occluding contour in a precise mathematical sense.
    \item[4] {\em Bumps are distinct parts} of a shape. As such, they are bounded from one another and should be manipulable separately. We introduce several novel visual illusions that illustrate this reversibly. 
    \item[5] {\em Bumps have both image and scene signatures.} To to be perceivable, there must be some image signature to the bump. To define this we build on a previous theory of shape, and define bumps using extremal curves of slant. We show invariance to many aspects of lighting and material changes, and provide evidence that this image signature is represented in visual (temporal) cortex. 
\end{itemize}

\begin{figure}
\includegraphics[width= 1 \linewidth]{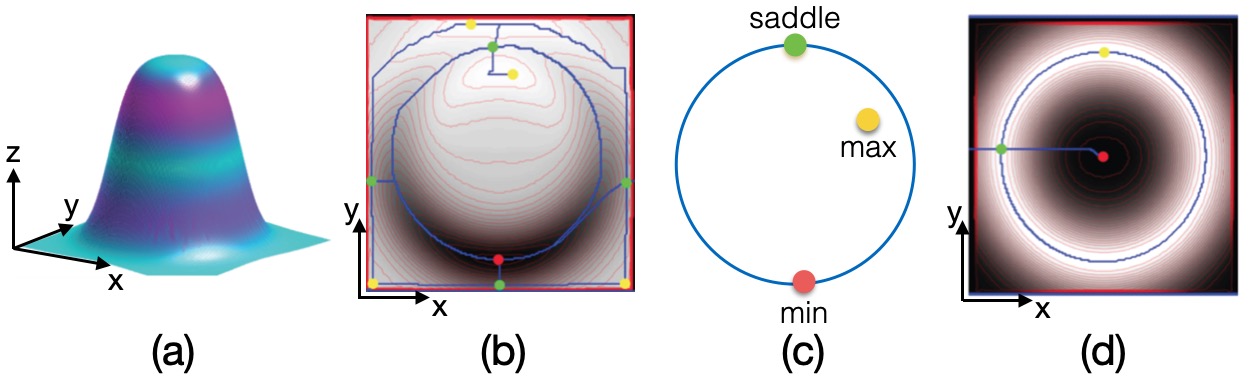}
\caption{The definition of a bump. (a) A bump in 3D, expressed as height function $z = z(x,y)$. (b) An image of the bump from above. The bump description is sketched by a circular contour (blue) that cuts through isophotes (green) and passes through critical points (minimum in red; saddle in green) of the intensity function. The interior maximum is due to the light source. The precise manner in which it cuts through the isophotes is key. (c) The blue curve is a closed extremal contour, here shown as an abstract graph. Together with the critical points (maximum, minimum, saddle), this template provides the definition of a bump. While the max can move around, due to lighting changes, it must remain in the interior. (d) The slant function of the surface in (a). Importantly, the same topological features (critical points and extremal contour) are it, establishing a relationship between the image domain and the scene domain.}
\label{fig:intro}
\end{figure}

A preview of our result in shown in Fig.~\ref{fig:intro}. Notice how there is a closed contour surrounding the bump, and that it passes through two critical points in the intensity function. It is helpful to think of this curve as a stroke drawn by an artist. Different artists might sketch it slightly differently; the commonality between them is actually part of a topological description of image structure -- a {\em critical contour} -- that was introduced in \cite{Kunsberg18}. It captures the global and qualitative aspects of shape, and will be reviewed in the Background section of this paper. This critical contour, together with the critical points, provides a template of our formal definition of a bump. It is expressed in terms of image properties, as was the development of critical contours. In this paper we work the other way around: we introduce the shape-domain counterpart to critical contours, which we call {\em extremal curves of slant} and show how they are related. This new derivation builds on the occluding contour, a more intuitive starting point, and highlights some of the subtlety in working with surfaces. In particular, we concentrate on generic surfaces built, viewed, and illuminated generically, exploiting the fact that small changes in the world should lead to small changes in (parts of) the image, almost always. We concentrate on closed extremal curves, because these outline bumps, and provide novel demonstrations of how bumps can be multi-stable even though the base of the shape remains fixed. While this confirms how bumps are distinct components of shapes, these components are not completely independent of one another -- they need to fit together like pieces of a puzzle -- even though their influence on the shape between them is much weaker.

The paper is organized as follows.
In the next section we develop the background necessary to explain our approach. We begin with a discussion of the occluding contour, how isophotes concentrate near it, and how the surface normal is known along it. This illustrates the 'holy grail' of shape inference -- a link between image and surface properties -- and it does so in terms of isophotes. But other aspects of shape perception are different: The occluding contour is precisely the loci of positions (on the object) where the view vector just grazes it; away from the occluding contour shape perception is more variable, subjective, and elusive. We illustrate, for example, how subjects agree on some aspects (e.g., the approximate location) of bumps, but differ on others (e.g., the slope and depth); others are reviewed below. We interpret this to imply that shape perception is qualitative and not quantitative, and therefore topological in nature. We review topological concept of the Morse-Smale complex informally, and sketch our previous theory of {\em critical contours}. With this foundation, we start the formal development of extremal contours. We work in the slant domain, specialize them to closed extremal contours, and provide examples of their computation on actual shapes. This abstract development is then brought back to ground by showing how closed extremal contours effect the isophotes, thereby providing a biological way to compute them. This has the added advantage of linking shading inferences to texture inferences. An application to neurobiological stimuli supports their relevance to primate vision, and an extension to specular imagery supports their universality. Theoretically we establish the equivalence between critical contours and extremal curves of slant for generic surfaces and rendering functions. In the end this connects the orientation basis for early vision to topological descriptions of surfaces, thus closing the loop between appearance (in the image) and (aspects of) shape in 3D space. That the full quantitative detail of the shape is not recovered is a consequence of the ill-posedness of the problem; that the qualitative detail of the shape is recoverable provides a foundation for human perception of shape.\footnote{An earlier version of this material was presented at \cite{zucker2019borders}.}

\section{Background}

Shape inference involves establishing a relationship between the image domain, on which our perceptual inferences are grounded, and the scene domain, on which the associated three-dimensional semantics exist. There are few examples where this relationship is clear and well founded; the occluding contour is perhaps the best known one. We review this material to motivate \emph{slant extremal contours},  which have special shape significance (due to their relationship with occluding contours) and special perceptual significance (due to their relationship with critical contours). In the process, we also review the notion of critical contours, illustrate their biological significance, and tie them to the qualitative nature of bump perception.

\subsection{Occluding Contour}
\label{sec:occl}

We now show how the occluding contour relates image and scene domains and implicates isophotes. We begin with standard definitions; see e.g. \cite{marr1982vision}.

The \emph{rim} of an object is composed of all non-interior points where the view vector `glances' the object; that is, the view vector lies in the tangent plane to the surface.  The \emph{occluding contour} is defined as the projection onto the image of the rim of the object. A powerful (but often elusive) cue, it has been studied in \cite{koenderink1984does, koenderink90Koen, Lawlor200918} among many others.

Two properties are key.  First, the occluding contour directly informs the viewer of the local surface normal; since the view vector lies in the tangent plane, it has \emph{surface meaning}.   Second, the occluding contour has a consistent flow `signature'; so it also has \emph{image salience}. We develop these in turn, starting with a standard representation for surfaces. 

\begin{figure}
\centering
\includegraphics[width= 0.4 \linewidth]{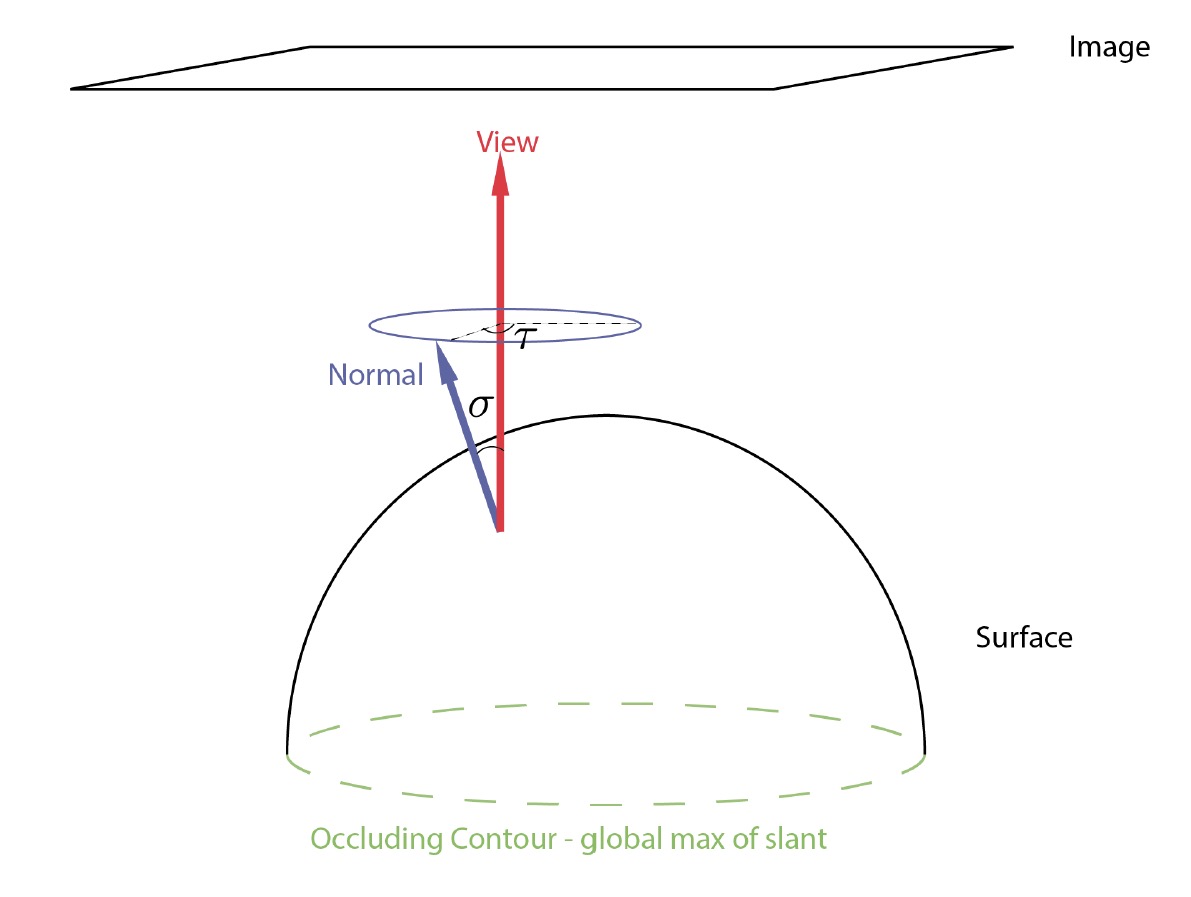}
\caption{The orientation of a normal vector to the surface at a point can be specified in slant $\sigma$ and tilt $\tau$ coordinates. Since there is a normal at every point, these coordinates define a scalar function over the object. Note how the occluding contour is given by maximal values of the slant.}
\label{fig:slant-tilt}
\end{figure}

The \emph{slant} $\sigma(x, y)$  is the polar angle between the surface normal and the view direction.  The \emph{tilt} $\tau(x, y)$ is the azimuthal angle between the surface normal and the view direction; see Fig.~\ref{fig:slant-tilt}. Both can be considered as scalar functions on the image domain.  Of course, these functions are unknown when the surface is unknown.

The surface normal and the view vector are perpendicular at every point on the rim, so the slant on the occluding contour is $\pi/2$. Thus, the occluding contour directly informs the viewer of the slant. This is a rare and unique property for a contour identifiable from the image. Importantly:

\begin{remark}{}
The slant achieves a global maximum on the occluding contour, since the slant of the visible surface must always be bounded by $\pi/2$.
\end{remark}

\begin{figure}
\includegraphics[width= 1 \linewidth]{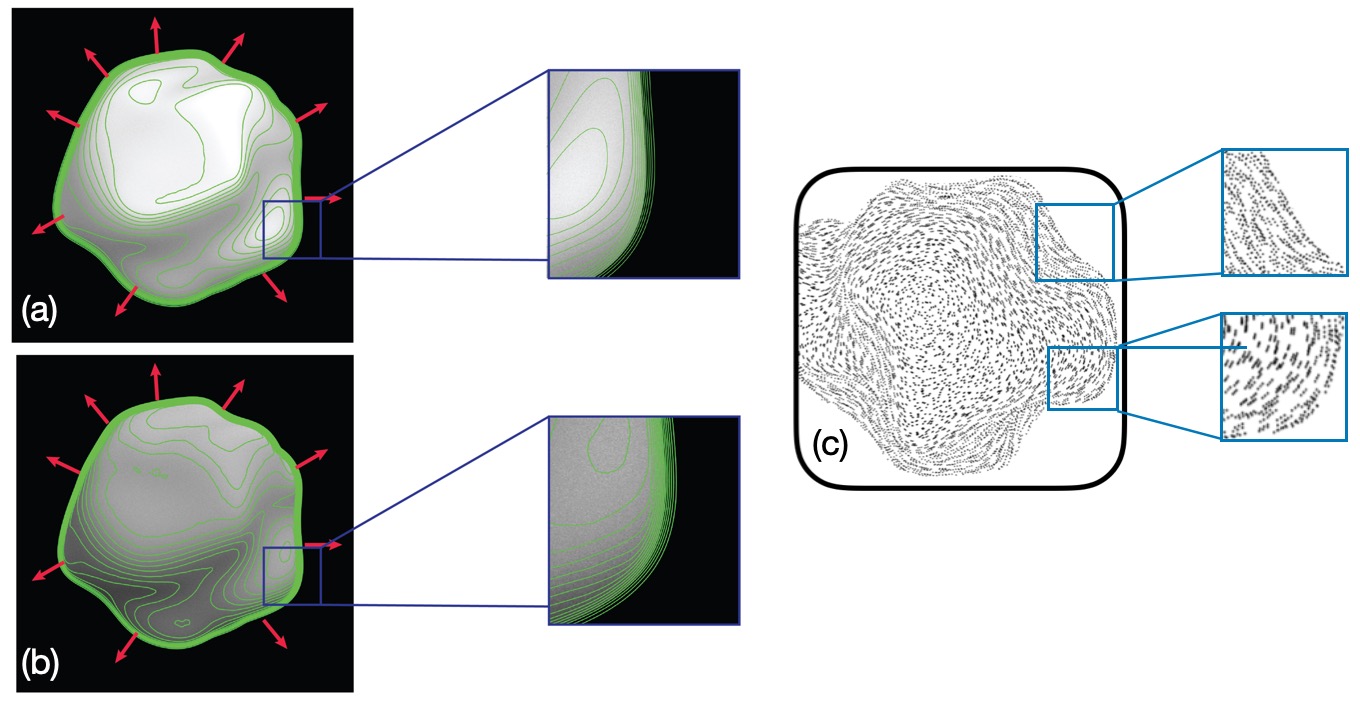}
\caption{Image salience, surface salience, and isophote concentration. 
    (a): A random shaded object with and associated isophotes in green.  Note how the isophotes concentrate near the occluding boundary (image salience) and how the surface normal field (red arrows) is known along the occluding contour (surface salience).  (b) A different rendering of the object. Note how the isophote direction remains parallel to the occluding contour regardless of the rendering function near the occluding contour. (c) A random object covered with a Glass pattern texture; again texture orientation becomes tangent to the occluding contour. Both shading and texture phenomena are due to the increasing projection near the occluding contour. We shall modify these effects to apply to bumps. (a, b) after \cite{Kunsberg2018Focus}; (c) after \cite{holtmann2012superposition}.  \label{fig:occluding}}
\end{figure}

The second property of the occluding contour is a common image signature due to foreshortening (Fig.~\ref{fig:occluding}).  Plotting the isophotes, or level sets of image intensities, reveals how they concentrate as the occluding boundary is approached. If the image were rendered smoothly from the surface, then the image flow would be tangential to the occluding contour as it is approached from the surface interior. One way of describing this is that the tangent plane to the surface "folds" away from the viewer as her glance approaches the rim. This condition has been studied computationally \cite{huggins2001folds, huggins2001finding} and psychophysically \cite{palmer2008extremal}.

Isophotes yield a smooth flow -- their tangent map -- when sampled by visual cortex  (\cite{ben2004geometrical, Kunsberg2018Focus}. Such fields of local orientations enjoy the same two key properties (Fig.~\ref{fig:occluding}(c)). We note, in particular, that Glass patterns emphasize this property, with interior bumps surrounded by an enclosing flow. We shall later exploit this property as a scheme to identify bumps in textures.

At the object rim, any vector component in the view direction will project to a zero vector in the image.  Thus any vector (brightness gradient, texture element) lying in the surface tangent plane will project tangentially to the occluding contour.  This means: (i) the orientation flow becomes tangent to the occluding contour on the inside and (ii) there is a discontinuity in the orientation flow on the rim due to the foreground/background transition; i.e., toward the outside.  We note that this property does not depend on the rendering and will call it {\em image salience}. 

These two properties of the occluding contour are special.  Because the occluding contour has an image signature regardless of the cue (brightness or texture), it can be identified from the image.  Because it has surface meaning, it constrains the surface.  This is an example of a direct map from an image property to a surface property with no ambiguity.  One of the main results of this paper is showing the existence of similar curves that have these same two properties (although slightly weakened) that are interior to the shape. Thus they are a natural generalization of the occluding contour.  

\begin{remark}{}
We conjecture that oriented flows in the image can surround bumps.
\end{remark}

\subsection{The Perception of Bumps}

\begin{figure}
\begin{center}
\includegraphics[width = 0.8 \linewidth]{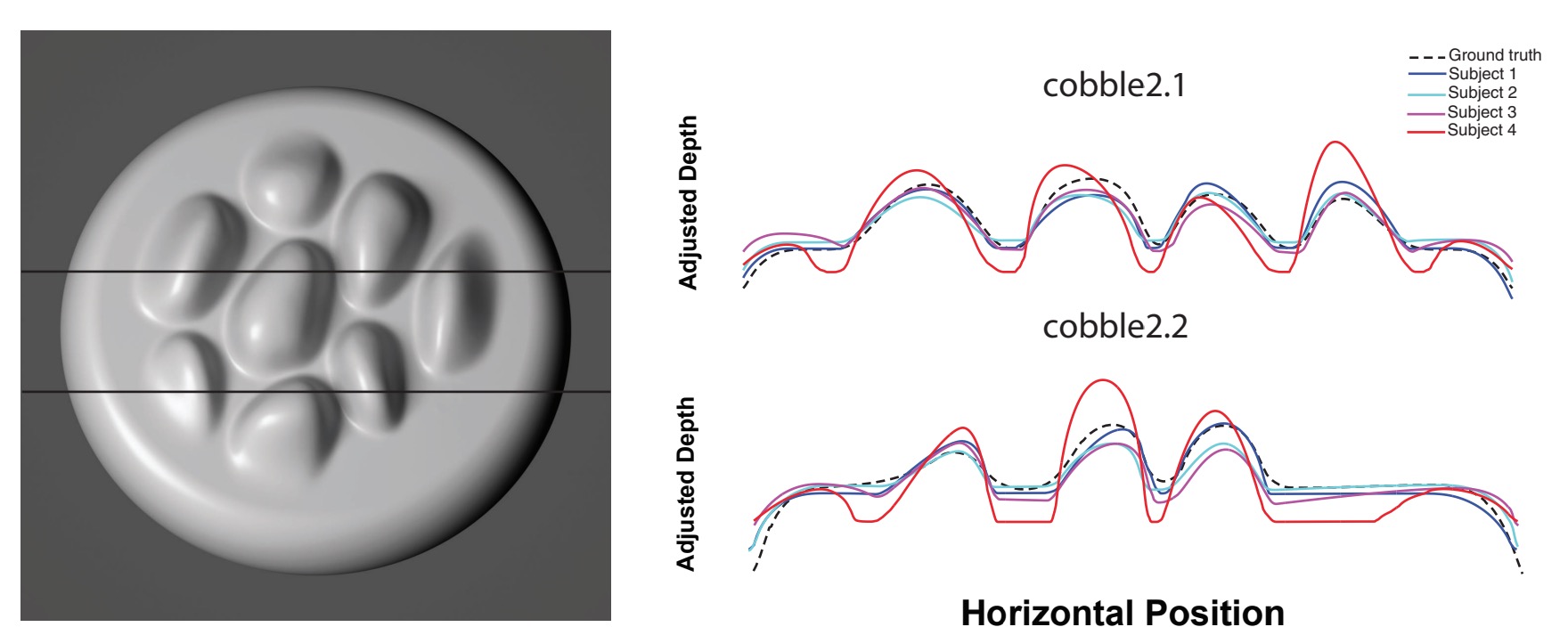}  \\
\caption{Bump perception is qualitative (left) An image showing a number of bumps on a surface. Superimposed are two scanlines, along which subjects estimated the height. The results (right) show that each subject perceived a slightly different profile, although there is substantial agreement in the general outline: we say that the heights and the boundaries are in qualitative, but not quantitative agreement.  Figure from \cite{ToddVSS17} \label{fig:todd}}
\end{center}
\end{figure}

We now turn to the psychophysics of bump perception, and highlight one study \cite{ToddVSS17}\footnote{We thank James Todd for access to data and figures.}. Subjects viewed a series of images of bumps on a background, rendered under different conditions (Figure~\ref{fig:todd}; see also Fig.~\ref{fig:cobblestones}). Their task was to estimate depth (the third dimension) along scan lines. While all  subjects were in agreement about the qualitative nature of the bumps, there was substantial quantitative variation across them (see Fig.~\ref{fig:todd} (right)). This qualitative variability contrasts sharply with the crisp identity just illustrated for occluding boundaries.

Such results are typical for shape perception, even with specular reflectance functions \cite{Mooney14, ToddVSS17}.  Both of these studies investigated the 3D perception of rendered random objects using either a scanline task or a gauge figure task.  Both results can be interpreted similarly.  They show that humans do not perceive veridical depths; the perceived heights of the `stones' in Fig.~\ref{fig:todd} are not only mostly wrong but also incoherent. However, the perceptions are quite accurate in the localization of the 'bumps' on the surface, e.g. the edges of the `stones' are accurately perceived, which is precisely what the extremal curves of slant capture.  Many more psychophysical studies \cite{doi:10.1167/12.1.12, MAMASSIAN19962351,Mingolla1986, doi:10.1068/i0645, doi:10.1167/15.2.24, CHRISTOU19971441, doi:10.1167/15.2.24, Seyama19983805,Curran19961399, doi:10.1068/p5807, doi:10.1068/p251009} show similar results.

\begin{remark}{}
Bump perception is founded on a qualitative description.
\end{remark}

\subsection{Critical Contours and Qualitative Shape Representations}

The qualitative nature of shape perception can be generalized from shading. A vivid 3D shape percept can be achieved from a masterful sketch, where the information is purely a set of lines and the corresponding white space.  How might the brain 'reconstruct' from this insufficient information?  It cannot hope to infer \emph{the} veridical perception; however, qualitative judgments such as relative depths can still be made in some circumstances \cite{koenderink15Koe}.  We submit \cite{Kunsberg18} that these two observations are linked, and that shape inferences from shading (and texture \cite{cholewiak2014predicting}) are, like shape inferences from contour, qualitative and relative. We have developed an approach for doing this, which we now informally introduce. This paper can be viewed as an application and extension of that theory, together with a very different motivation; further details and references in \cite{Kunsberg18}.

'Quantitative', as the term is typically used, connotes accurate, numerical and formal, while 'qualitative' can suggest vague and 'informal.' We use the term qualitative in a formal, mathematical sense, to denote topological rather than geometrical ideas. Topology studies bumps and valleys; differential geometry studies tangents and curvatures. Topology is global (invariance over rubber sheet deformations), differential geometry is local (curvature at a point). We shall exploit this idea of smooth deformations when we define extremal contours.

\begin{figure}
\centering
\includegraphics[width= .8 \linewidth]{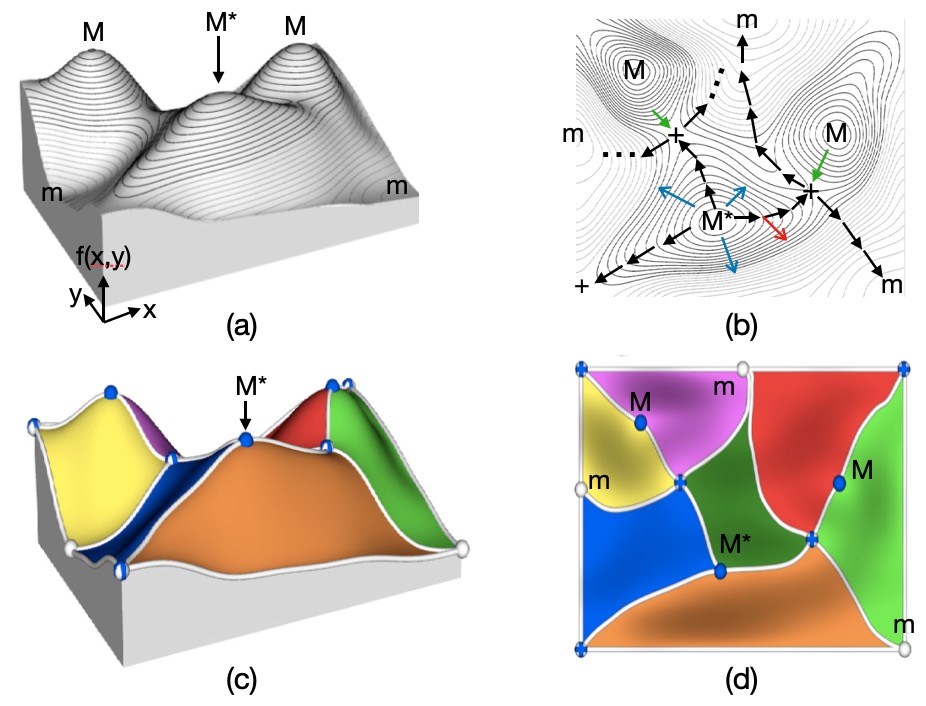} \\
\caption{Brief introduction to the Morse-Smale complex. (a) Consider a scalar function $f(x,y)$ as a smooth mountain range (top view in (b)). Note how the level sets are nested and circle the maxima. (b) Imagine pouring water on the central peak (M*); most will flow downhill to nearby minima m along the gradient (blue arrows), cutting across the level sets. A special flow would follow distinguished paths along ridge lines (black arrows) from the maxima to saddle points (+) and then to minima. Should the flow deviate from the path, it will fall directly toward a minimum (red arrow). Other maxima may flow into these saddles (green arrows) so that in general, for each saddle, there are two entering and two leaving paths. (c,d) The Morse-Smale complex is a graph, in which the nodes (0-cells) are extrema (maxima, minima, and saddles) and the edges (1-cells) connect maxima to saddles and saddles to minima, thereby linking the black arrows. The graph forms quadrilaterals, called 2-cells, that provide a tesselation of the domain into components. Figure modified from \cite{Gyulassy08}; see further discussion in \cite{Kunsberg18, Kunsberg2018Focus} and the Appendix.}
\label{fig:guylassy}
\end{figure}

James Clerk Maxwell \cite{maxwell1870hills} started the topological description of landscapes by examining how water flows downhill from mountaintops to valleys, along ridge lines and courses, and settles in minima.  In brief, water flows down the gradient to minima, except when it follows a ridge line exactly and flows from a maximum to a saddle (Fig.~\ref{fig:guylassy}(a,b)). These ridge lines are special because they separate the flows toward one minimum from those toward another; any deviation off the ridge line and the water is drawn toward a minimum rather than the saddle. Of course, once the saddle is encountered the flow is then toward a minimum. Note, in particular, that the flow is along the gradient of the surface; this is perpendicular to the level lines everywhere, and that it would remain relatively unchanged (i.e., qualitatively similiar) if the $(x,y)$ domain were smoothly stretched or compressed. In a sense one can think of generic landscapes as those in which the flows follow these observations, without passing through constant planar regions at intermediate positions. 

The modern form of these observations is elegantly captured in the Morse-Smale complex, a rigorous way to qualitatively describe a scalar field via a set of curves \cite{Biasotti:2008:DSG:1391729.1391731, smale1961gradient}; see Fig.~\ref{fig:guylassy}(c,d). It results, in effect, in an abstract, graphical version of what was just described. The mountain range becomes the value of a scalar function $f(x,y)$ which could be image intensity or surface slant. The nodes of the graph ({\em 0-cells}) are the extrema of this scalar function, or places where its derivative is 0; edges in the graph ({\em 1-cells}) connect maxima to saddles and saddles to minima. Notice, in particular, how cycles of four edges ({\em 2-cells}) are formed, connecting a maximum, a minimum, and two saddles in alternating order. These 2-cell quadrilaterals segment the mountain range into characteristic domains. We shall shortly be modifying these components to develop the abstract definition of a bump as a special type of domain.

\begin{remark}{}
The Morse-Smale complex is a topological description of a function; it makes certain of its shape features explicit but does not specify the precise function values everywhere. Values on the complex can be used to get a `weak' representation of the original function.
\end{remark}

We use the Morse-Smale complex (MS) as the foundation for our shape description.
Just as a sketch is a union of thin black lines on white paper, {\em critical contours} \cite{Kunsberg18} are a concentration of shading. Imagine a drawing of a ridge: a thin line, perhaps drawn by an artist, would be the limiting case in which the critical contour has infinitely-steep intensity 'walls' surrounding it (see Fig.~7 in \cite{Kunsberg2018Focus}). Informally, critical contours can be viewed as a sketch of the shading inside a shape, just as an occluding contour is a sketch of the boundary of a shape. Formally, the critical contours are those edges (technically, 1-cells) of the Morse-Smale complex that have large gradients surrounding them.  Critical contours are computable from the image, and a main result of that theory is critical contours are nearly invariant to changes in the rendering function.  In effect they define a type of scaffold on which a shape can be built.

\begin{figure}
\centering
\includegraphics[width= .7 \linewidth]{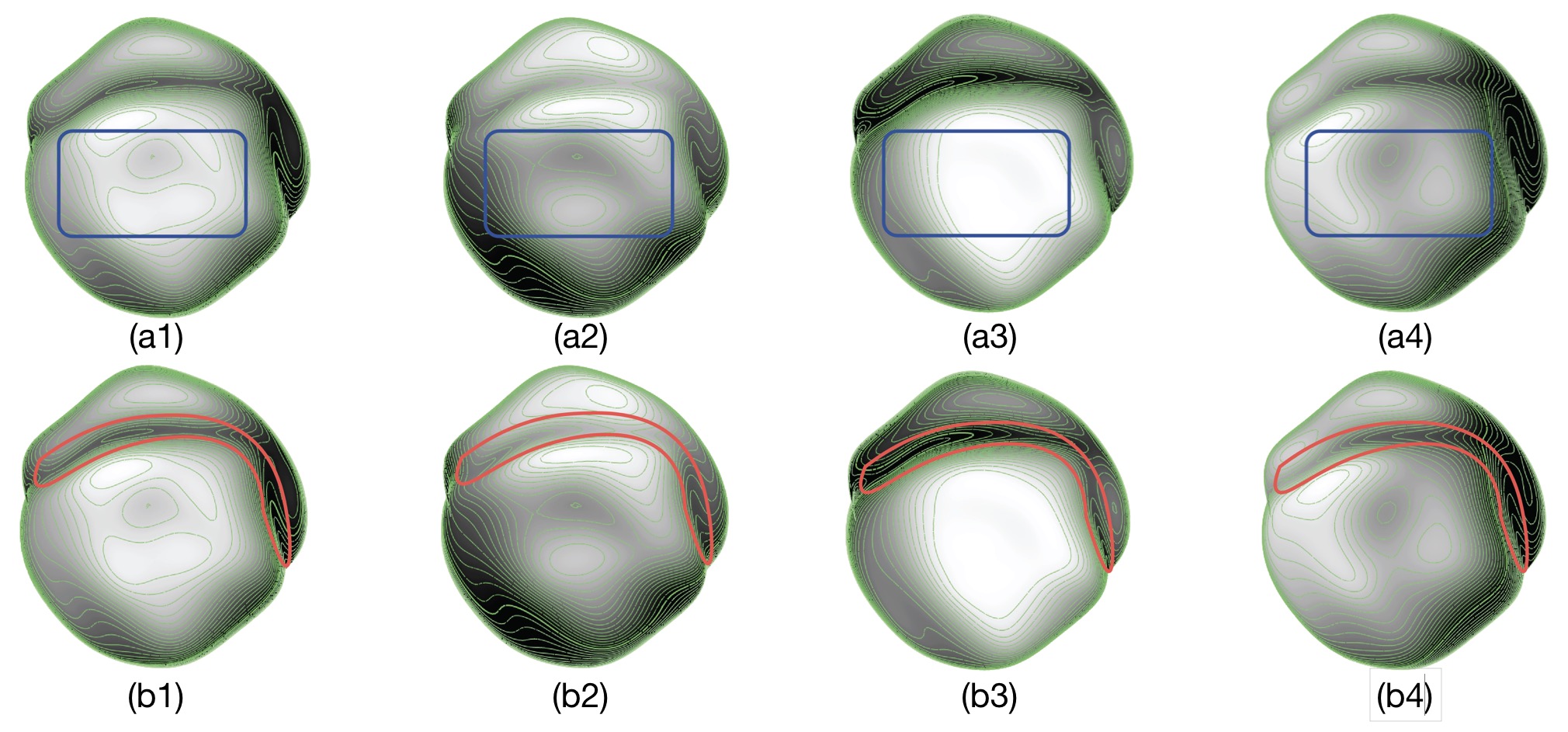} \includegraphics[width= .2 \linewidth]{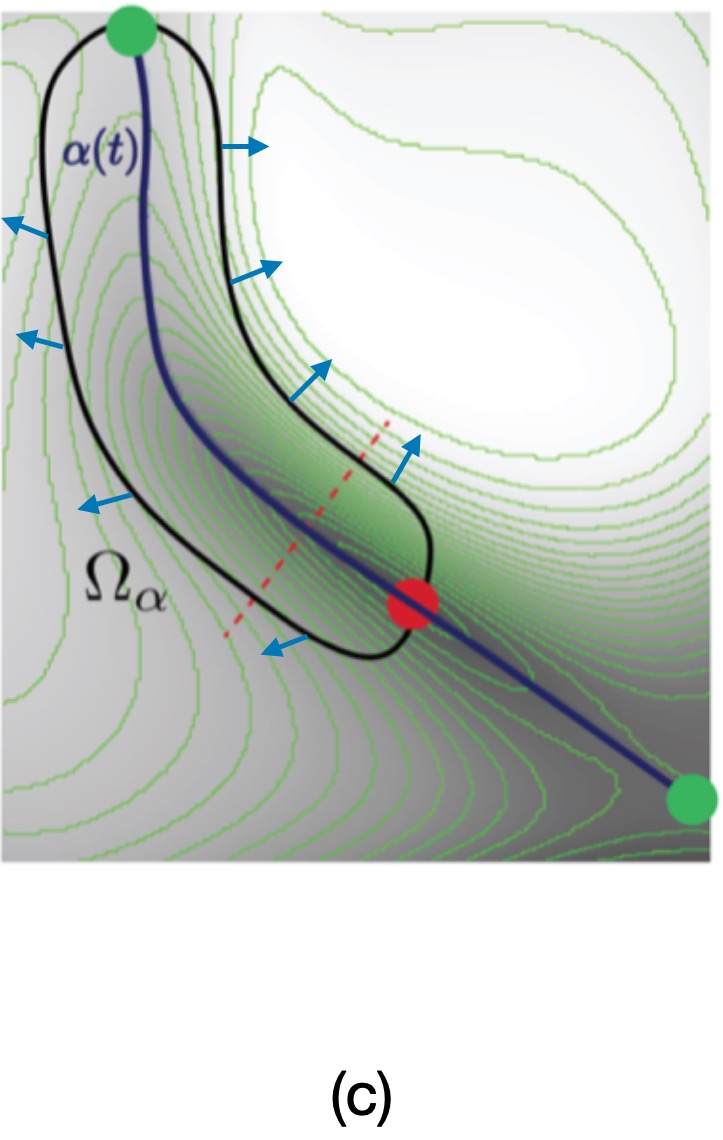} 
\caption{Isophotes, or their local orientation (the shading flow) abstract the cortical representation of smooth intensity distributions. For a random smooth object they change significantly with lighting in most places (e.g., within the boxes in (a)), they are essentially invariant within the red sausages in (b). (c) Critical contours (the blue curve) capture this invariance. Notice how they cut across the nested isophotes around a ridge, and in this instance pass from a saddle (green) to a minimum (red). The dotted red line shows that the intensity variation across the critical contour goes rapidly from bright to dark to bright, with the gradient (blue vectors) pointing away from the critical contour on both sides. (c) after \cite{Kunsberg18}. }
\label{fig:moving-isophotes}
\end{figure}

The 1-cells of the MS complex lie along the gradient flow, which is orthogonal to the level sets everywhere. A delicacy arises because, although many have suggested that the shading flow (along level sets) is the foundation for shading analysis (\cite{koenderink:1980bm, zucker-sff}), others have observed that the level sets change drastically with changes in lighting or reflectance (e.g., \cite{doi:10.1068/i0645}). While this observation is true in many places, it is not true in a neighborhood around critical contours (Fig.~\ref{fig:moving-isophotes}). Secondly, across neighborhoods like this the intensities change rapidly (the image gradient is large), from dark along the critical contour to bright in either direction normal to it. These two observations illustrate the basis for critical contours, and they hold generically for a wide class of intensity and surface variations \cite{Kunsberg18}.

\begin{figure}[ht]
\begin{center}
\includegraphics[width=0.8 \linewidth]{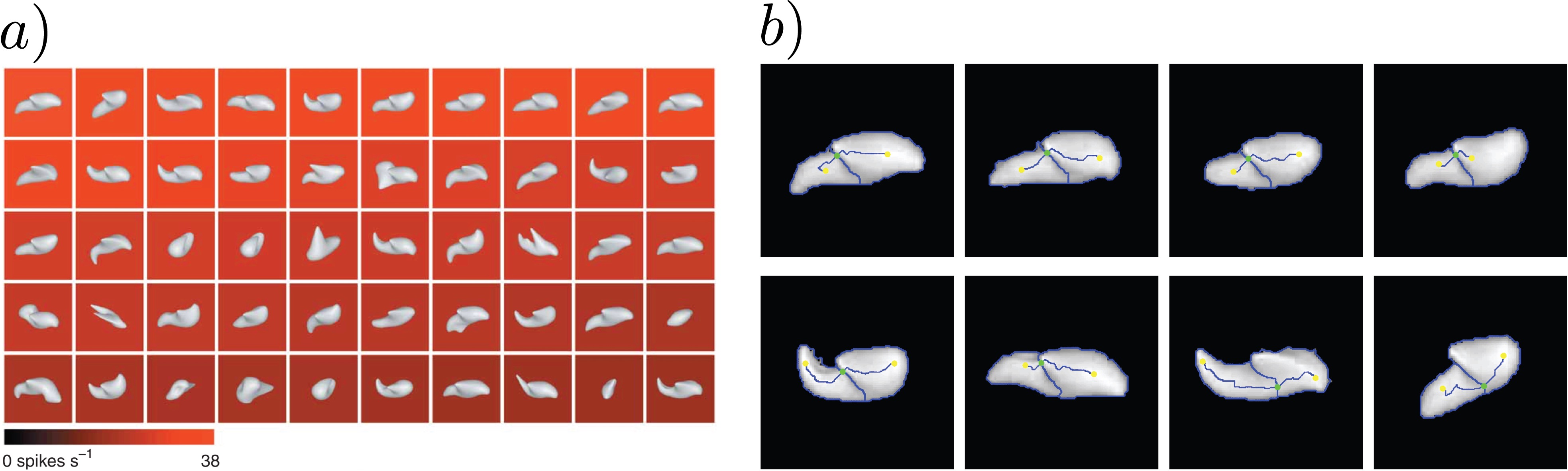} \\
\caption{\label{fig:yamane_results} Physiological evidence for critical contours. (a) A varied set of 3D stimuli are preferred by a neuron in macaque IT cortex \cite{Connor08}. Note the variability in shape and pose. (b) Even though there is substantial variation in shape and pose, the critical contours of the first 8 examples are topologically equivalent for the different stimuli.}
\end{center}
\end{figure}

In addition to the formal motivation, we now introduce biological support for critical contours. Yamane et al. \cite{Connor08}\footnote{We thank C. Connor for the use of these figures.} developed images of shapes that, despite shape and lighting variations, yielded consistent neural responses in visual area IT Fig.~\ref{fig:yamane_results}). The link tying these different cases together is a common critical contour.

\begin{remark}
Critical contours provide a topological signature of key interior shape components, stable under generic lighting and rendering variations.
\end{remark}

\section{Extremal Curves of Slant}
\label{sec:slant_extremal}

We now begin the technical contributions in this paper. Extremal curves of slant are motived from the rim, complementing our previous development \cite{Kunsberg18}, which started with the image.

\begin{figure}
\includegraphics[width= 1 \linewidth]{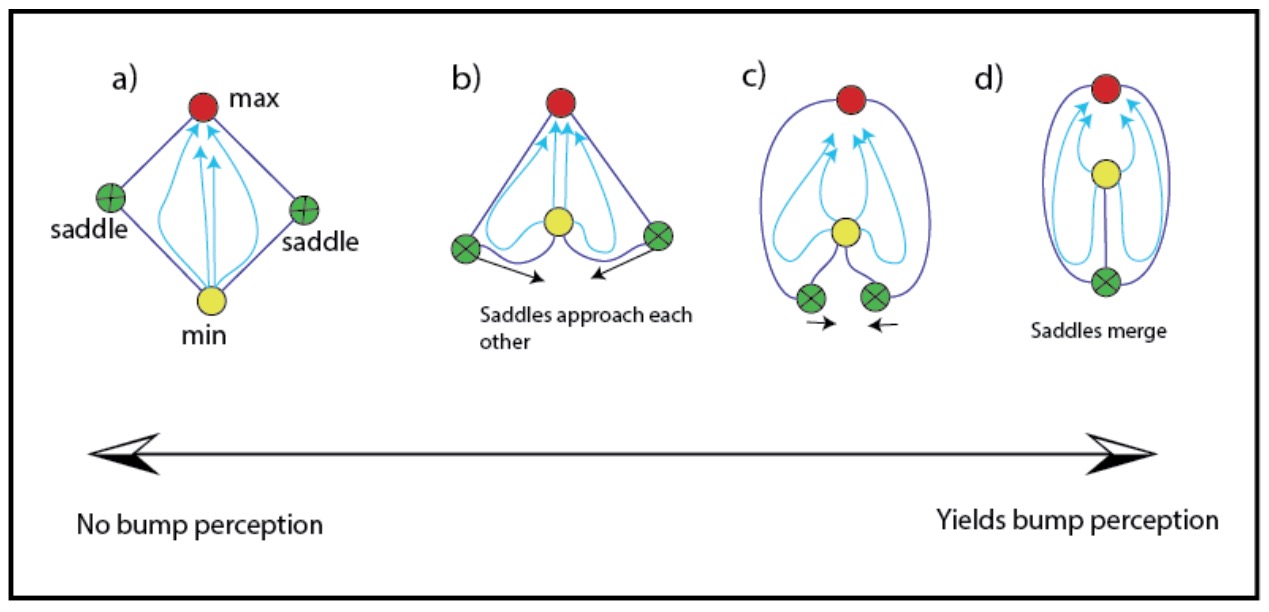}
\caption{Development of the MS-complex for bumps. (a) Starting with the standard MS cycle, (b) smoothly deform the domain so the saddles begin to (c) approach one another until (d) they merge. Notice how the integral curves (cyan) deform along with the saddles.}
\label{fig:saddle-merge}
\end{figure}

The first step is to "move" the rim into the interior of the shape, so that it could surround a bump. This requires a relaxation of the slant along the occluding contour to something less than global maxima at every point, since the tangent plane never rotates far enough to include the line-of-sight for a smooth bump. One might naively try to relax from the global maxima to some version of local maxima of slant along a contour, but this is impossible for technical reasons. The surface has to be generic. For Morse functions this requires critical points to be non-degenerate (their Hessian, or second derivatives must be full rank), and for the surface to be smoothly undulating. Small variations to the undulations should have small effects and "water" must be able flow smoothly unless it encounters a critical point. Mathematically, for a generic Morse function, there are no appropriate contours consisting entirely of local maxima of slant. 

There is a modification of the general surface shown in Fig.~\ref{fig:guylassy} that does work. The basic idea is not that slant is critical everywhere along the curve, but that is large along the contour and flows (along the gradient) from one critical point to another. This is why we call it an extremal curve of slant. In effect, we are seeking a curve around a bump that is steep everywhere, and such that every step across it is a large step toward the top. 

To develop extremal curves of slant, recall (Fig.~\ref{fig:guylassy}) that the 2-cell, the basic building block of the MS-complex, consists of a region surrounded by a special flow from a maximum to a pair of saddles and then to a minimum. While all flows are along the gradient, the 1-cells (the edges of the 2 cells) are special -- they delimit the building blocks with the flows into saddles. 
The appropriate generalization from the occluding contour to an interior curve, then, are the contours following the gradient flow that go through local maxima and saddles, but with a modification of the MS-complex in which the saddles have been joined. To derive these (Fig.~\ref{fig:saddle-merge}), begin with the standard 2-cell from the MS-complex for a scalar function, and deform it by allowing the saddles to approach one another until they merge.  This configuration is also generic, has been considered in the computer vision literature \cite{lee1984two, griffin1995superficial}. It gives rise to a distinct type of 2-cell typical for a bump.

\begin{figure}
\includegraphics[width= 1 \linewidth]{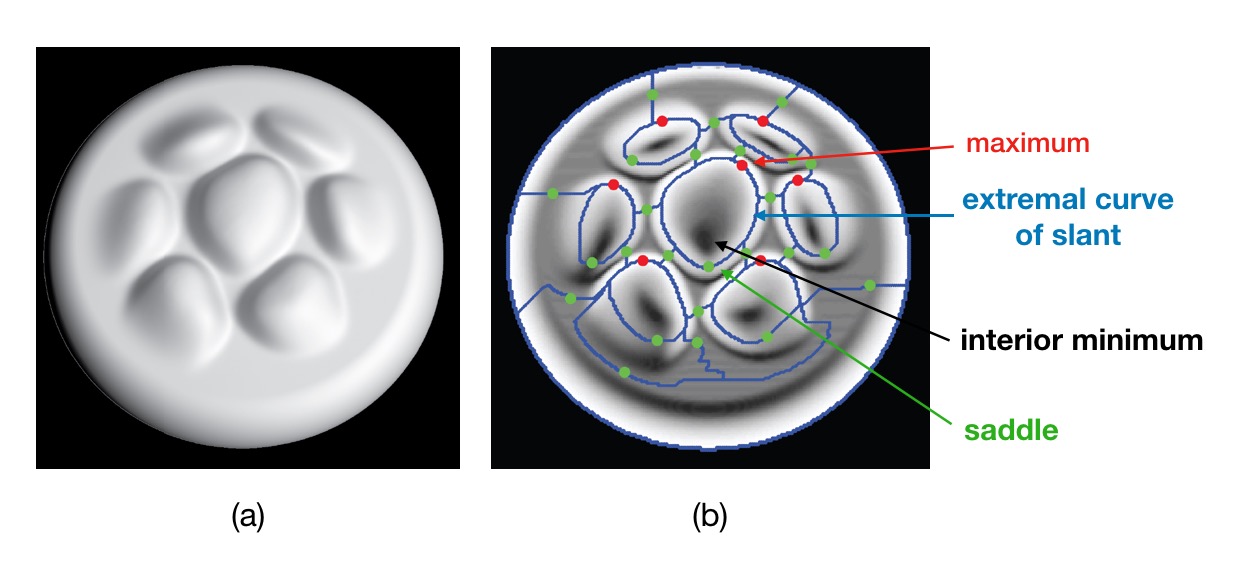}
\caption{Illustration of a bump max, saddle, and unknown interior min on a Todd image. (a) Intensity image. (b) Slant function for the surface. Typically the slant function has a minimum somewhere in the interior, while the intensity function often has a corresponding maximum (e.g., for Lambertian renderings). Notice how the extremal curve passes through the bright region denoting high values of slant; the portions of the MS-complex other than the closed extremal curves are not critical contours. \label{first-bump}}
\end{figure}

Our first example is an image from the Todd database (Fig.~\ref{first-bump}). Begining with the slant function,  notice how each bump is surrounded by an extremal curve that passes through a maximum and a saddle; this maximum is the 'steepest' part of the bump. Interior to the bump is a minimum in slant that often corresponds to a maximum in intensity, e.g.for a Lambertian surface illuminated near the observer. Note that the extremal curve clings to the high-slant portion of the function, and that this bright region (high slant) encloses -- and is surrounded by -- darker, low-slant neighborhoods.

In multi-bump images such as these, the MS-complex provides some constraint on the "space" between bumps. This space typically undulates slowly and provides a number of saddle points that can connect the bumps together. Unlike the extremal curves, these 1-cell/saddle connectors are not stable; they can move a good distance with changes in lighting or material.

\begin{figure}[h!]
\begin{center}
\includegraphics[width= 0.4 \linewidth]{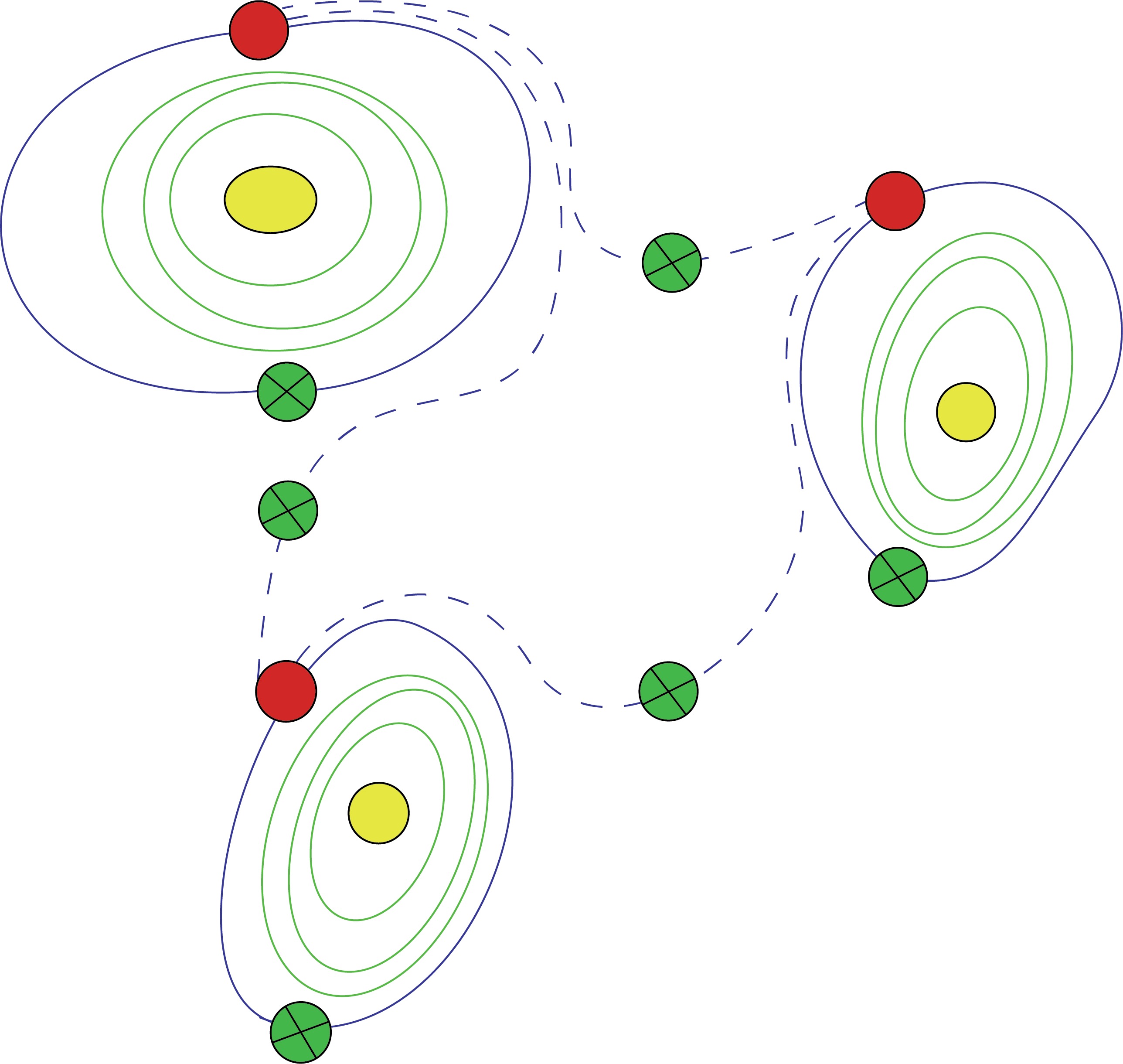}
\caption{The topological network (part of an MS-comples) corresponding to several bumps shown in a "cobblestone image" arrangement. The saddles link maxima of slant on nearby bumps together, although they are not as stable as the extremal curves surrounding each bump. The iso-slant contours surrounding each bump minimum will be discussed shortly.}
\end{center}
\end{figure}

\begin{figure}
\includegraphics[width= 1 \linewidth]{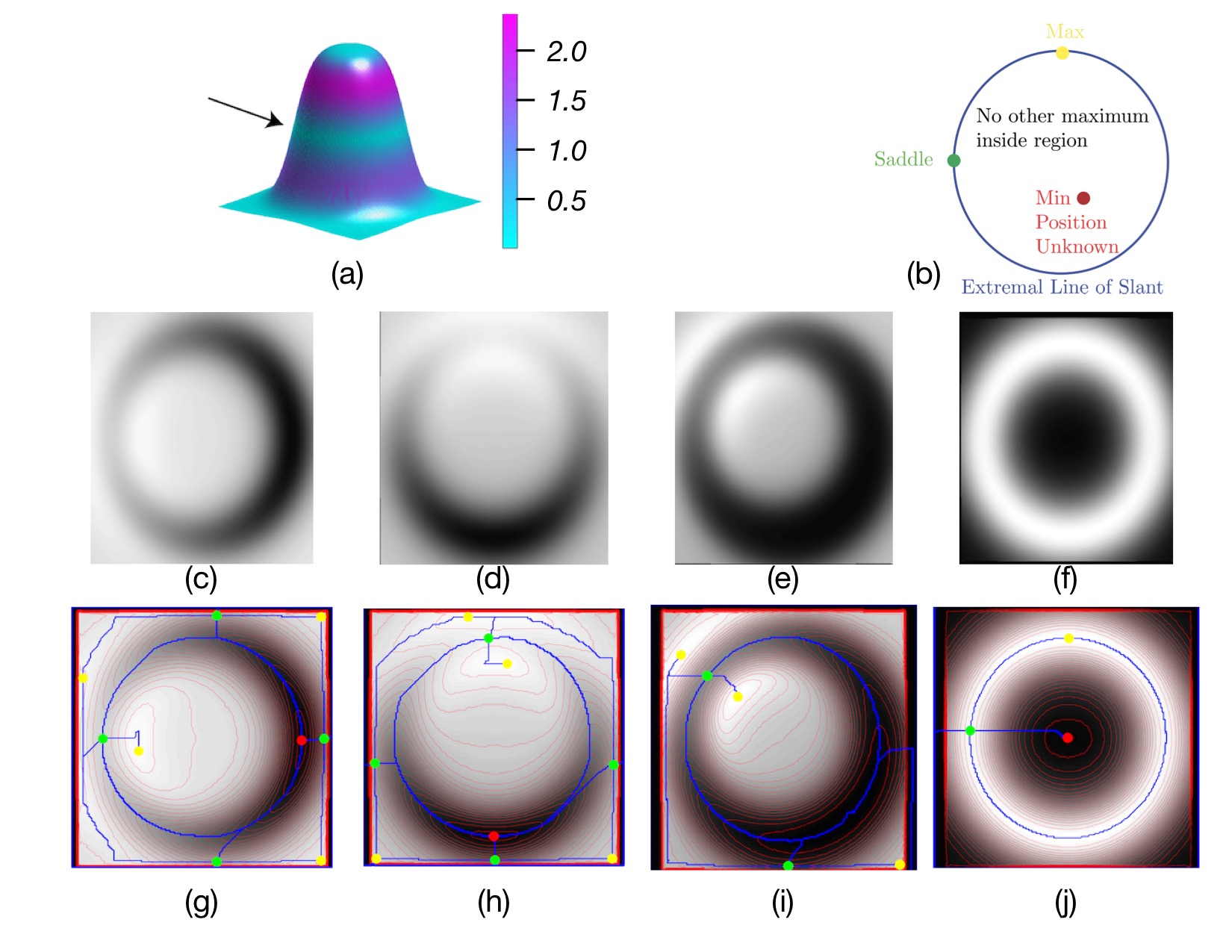}
\caption{ A sigmoidal bump on a slightly bent surface. (a) The bump is colored by the Gaussian curvature; the arrow points to the parabolic curve along which the Gaussian curvature is zero. (b) The defining template for a bump, shown in the slant domain so that an extremal curve surrounds a minimum. (c, d) A Lambertian reflectance function on the bump illuminated from two different positions. (e) A specular rendering function. (f) The slant function for the bump viewed from above. (g - i) The isophotes and MS-complex for the three images and (j) for the slant function. 
\label{fig:sigmoid}}
\end{figure}

It is helpful to analyze an artificial image of a sigmoidal bump (Fig.~\ref{fig:sigmoid}) in more detail. Starting from the top, notice how the slant is minimal, then increases to its maximum and then decreases again. While this maximum is not $\pi/2$, it does illustrate how the extremal curve is a relaxation of this aspect of the occluding contour. The MS-complex on the slant function shows how the extremal contour encircles the bump, with a slant minimum (and no other maximum) inside it. Also shown are three different renderings of the sigmoidal bump, each of which has a different intensity distribution and, necessarily, different isophote arrangements. Notice how the (image) extremal contour cuts through these along a distinguished path -- it is this path that remains nearly invariant over lighting and rendering changes. 

\begin{definition}{}
Slant extremal contours are the saddle-maxima 1-cells of the MS complex of the slant function.
\end{definition}

Slant extremal contours are interior contours, defined on the surface, that pass through the maxima of the slant function, follow the gradient flow, and, at the same time, segment the surface into regions.  For brevity in the remainder of this paper, we will drop the word `slant' and call them simply `extremal contours'.

In general, extremal contours will not be closed.  However, motivated by the previous examples we are especially interested in the special case when an extremal contour connects back to itself.

\begin{definition}{}
Extremal rings are closed slant extremal contours.
\end{definition}

Putting these pieces together, we have (up to a concave/convex reversal):

\begin{definition}{}


A bump/valley is an interior region within an extremal ring passing through a maximum and a saddle that enclose an interior minimum.

\end{definition}

\section{Extremal Rings and Occluding Contours}

 We now investigate extremal contours and extremal rings and show that these features define the surface topologically while also being visually salient for many rendering functions. It will formalize the previous observations. In particular, we show that, in all likelihood, these extremal contours have the same two properties that the occluding contour had: \emph{surface meaning} and \emph{image salience}, provided the surface and rendering are generic.

\begin{figure}[h]
\begin{center}
\includegraphics[trim={0cm 5cm 7cm 5cm},clip, width = 1 \linewidth]{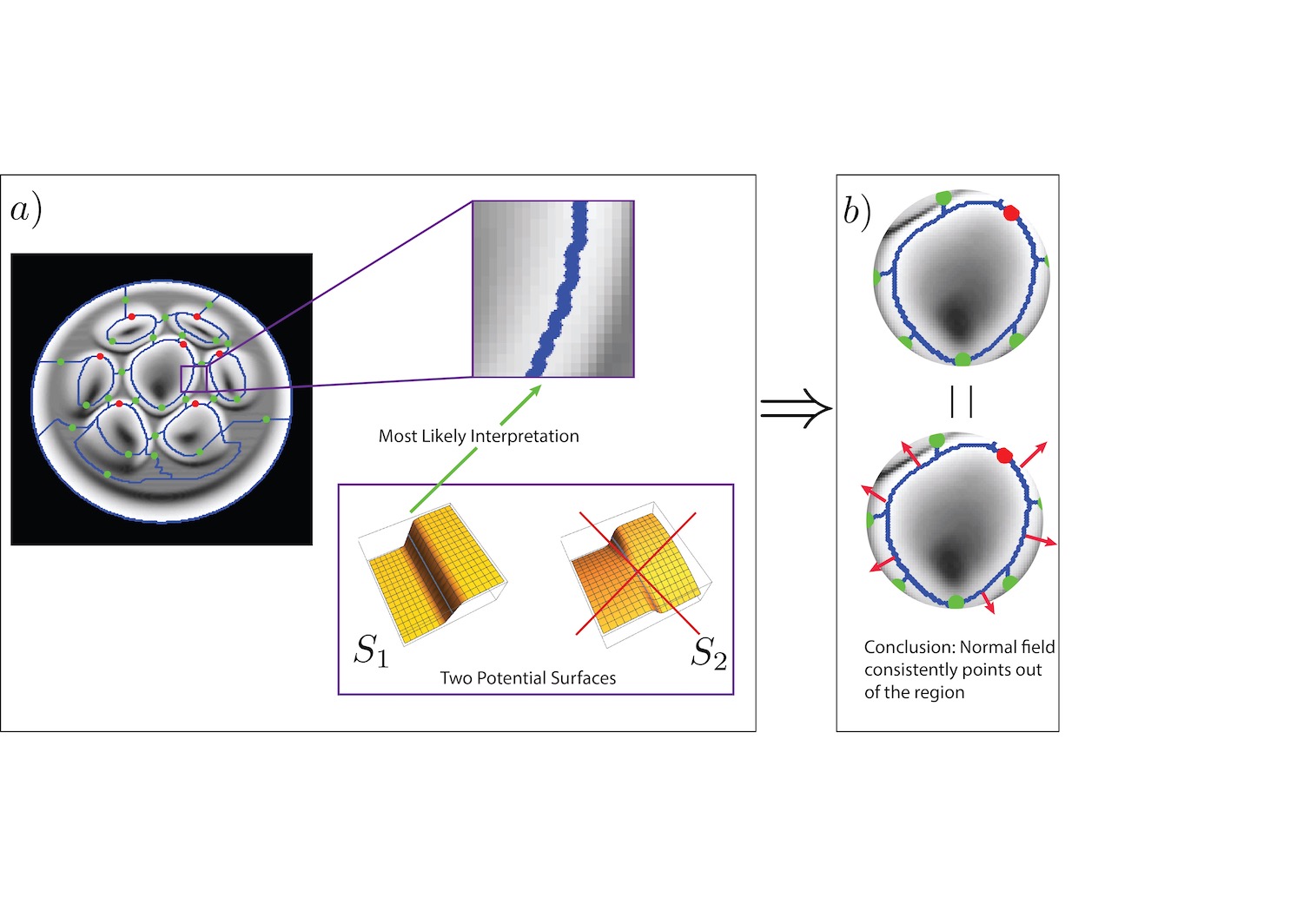}
\caption{Extremal curves and surface normals. (a): Consider a portion of a maximal curve of slant.  Two distinct types of local surfaces could have caused this local slant function, one with no twist and another with substantial twist.  The first surface, without twist, is much more likely than the second surface.  Note the normal field for the first surface consistently points to the same side (left) of the extremal curve.  (b) Taking the most likely interpretation for each portion of the extremal curve, the normal field must point uniformly outside (or inside) the entire contour. \label{fig:bdy}}
\end{center}
\end{figure}

\begin{figure}[h]
\begin{center}
\includegraphics[trim={0cm 5cm 0cm 3.2cm},width = 1 \linewidth]{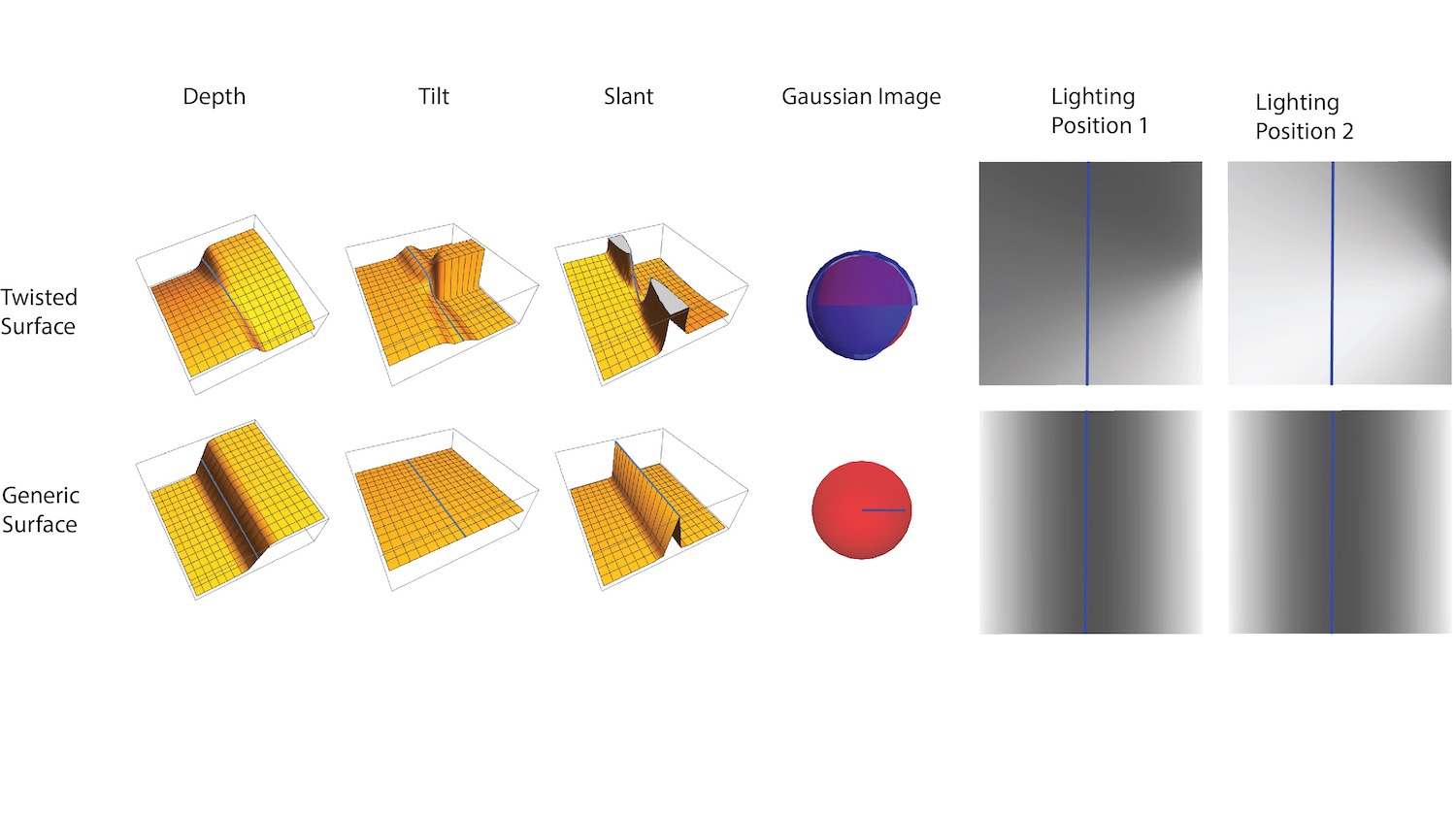}
\caption{Comparing possible surface explanations for a maximal slant curve. Each row depicts different properties of the surfaces shown in the first column.  Although both are technically solutions, the twisted surface covers most of the Gauss map and, when rendered under slightly different lightings, gives rise to drastically different images.  The generic surface solution occupies a small portion of the Gauss map and renders almost the same image under different light sources. Both arguments show that the twisted surface (top row) is much less likely. \label{fig:two_surfs}}
\end{center}
\end{figure}

\subsection{Extremal Contours: Surface Meaning}
\label{sec:ELSM}

We first show that extremal contours represent boundaries of bumps and valleys of the surface.  The argument will be based on the conclusion that the surface normal field should generically point uniformly to the interior (or exterior) of the region bounded by the extremal contours.  Our argument here is based on a generic prior similar to \cite{freeman94}.

Consider Fig \ref{fig:bdy}, and focus on the enlarged portion of the extremal ring. This was defined to cut across the level sets of the slant function but, by definition, it does not inform the tilt function. Here we show that, for generic surfaces, the tilt must be constrained along the extremal ring to prevent self occlusion and image instability. There are two basic possibilities: in the first case, the surface behaves like a bump boundary, where the surface normal on the extremal ring points consistently; the second case leads to rather wild surfaces with `crazy' curvatures. While this second case is a mathematical possibility (developed next), it violates our requirement that the surface be generic. That is, a small change in the lighting would lead to a drastically different image (Fig.~\ref{fig:two_surfs}). For the wild surfaces, the normal points in many different directions, covering a very large portion of the Gauss map (see additional information in Appendix).

More technically, let $\alpha(t)$ denote an extremal ring that bounds a region $R$, and let $\{x, y\}$ be image coordinates under orthographic projection.  Let $\sigma(x, y)$ represent the slant in a neighborhood of a point $(x_0, y_0)$ on $\alpha(t)$.  Rotate the frame so that the slant gradient (tangent direction to $\alpha(t)$) points locally in the $y$ direction.  We compare two possible solutions for the local surface depth $S(x, y)$ by defining two pairs of Taylor expansions for the slant and tilt functions.  Let $\sigma_1(x, y) = \sigma_2(x, y) = c_1 + c_2 y + \sigma_{xx} x^2 + \sigma_{xy} x y + \sigma_{yy} y^2$.  (There is no linear $x$ term here due to the slant gradient pointing along the $y$ axis.)
Let $\tau_1 = x$ and $\tau_2 = y$.  Now, $\{\sigma_1, \tau_1 \}$ defines a surface $S_1$ and $\{\sigma_2, \tau_2 \}$ defines an alternative surface $S_2$.  Which of these is more likely? 

Both $S_1$ and $S_2$ have the same magnitude tilt gradient.  However, $S_1$ (generic surface) has the tilt change along the contour while $S_2$ (twisted surface) has a tilt change perpendicular to the extremal contour. 

Let $N_1(x, y), N_2(x, y)$ represent the normal fields for each of these two solutions.  We compare the relative probability of each of these surfaces by considering the term $T_i = \int_\Omega \det(D N_i D N_i^T)$ for each surface.  This is essentially the Gaussian curvature of each solution integrated over the patch.\footnote{For an explanation of these terms, see \cite{holtmann2018tensors}.}  The solution with higher Gaussian curvature will be the solution that is less smooth, more dependent on lighting direction \cite{freeman94} and with a higher chance of occlusion.  We compare the relative likelihood of the two solutions $\frac{L_1}{L_2}$ by considering the inverse of the ratio $\frac{T_2}{T_1}$.  A simple calculation shows:

\begin{align}
\frac{L_1}{L_2} \propto
\frac{T_2}{T_1} \propto \frac{\sigma_{xx}^2}{\sigma_y^2}
\end{align}

where $\sigma_{xx}$ is the transversal second derivative of the slant across $\alpha(t)$ while $\sigma_y$ is the gradient along $\alpha(t)$.  Since the slant on $\alpha$ is extremal, its gradient will necessarily be small.  In addition, since the slant is changing rapidly across $\alpha$, $\sigma_{xx}$ will be large.  Thus, for a slant patch as shown in Fig \ref{fig:bdy}(a), the ratio $\frac{\sigma_{xx}^2}{\sigma_y^2}$ will be large.  This statement is illustrated by the comparison in Fig \ref{fig:two_surfs}.  We conclude that the image patch has a surface normal field that is not twisted.  In other words, it is most likely that the normal points to a single side of the curve in the entire Taylor expansion.

Applying the above argument completely around the extremal ring $\alpha(t)$ shows that the most probable interpretation of the surface normal field along $\alpha(t)$ is that it does not have a twist and therefore must point uniformly outside (or inside) the region bounded by $\alpha(t)$.  It then follows that the region $R$ is `higher' (or `lower') than the surrounding area.  More precisely, one can see that the $R$ must be an ascending or descending manifold of depth; in other words the region is a bump or a valley.

\begin{remark}{}
The surface normal along an extremal contour points consistently to the interior or exterior.
\end{remark}

\subsection{Extremal Contours: Image Salience}
\label{sec:invariance}
We now show that the second important property of the occluding contour, image salience, also exists for extremal contours.  Since extremal contours are related to critical contours \cite{Kunsberg18}, image salience will follow from the image invariance of critical contours, which we now review.


A given surface, when rendered differently (e.g. with a different light source), can yield drastically different images overall. This was illustrated in Fig.~\ref{fig:moving-isophotes}. However, many rendering functions such as Lambertian shading, specular shading, texture, Glass patterns, as well as line drawing algorithms \cite{DeCarlo03, Judd07} all involve the surface normal field in order to create an image. 

The surface normal field $N(x, y)$ can be defined as a map from the image domain $\mathbb{R}^2$ to the unit sphere $S^2$.  It is then natural to define a rendering function $F$ as a smooth map from the unit sphere to the real line, that is $F: S^2 \rightarrow \mathbb{R}$.  The image is then expressed as the combined map $I(x, y) = F(N(x, y))$.  The orientation field (for a smooth rendering function) is then computed perpendicular to the gradients $\nabla I$.

In \cite{Kunsberg18}, we defined critical contours that were computable from the image.  Generally,

\begin{definition}{}
Critical contours are gradient flows in the image with large transversal second derivatives. 
\end{definition}

\begin{remark}{}
Critical contours are computable from the image gradients whereas extremal contours are computable from the slant gradients.
\end{remark}

 In \cite{Kunsberg18}, we showed critical contours had an invariance to the choice of $F$.  More precisely, to show image invariance of a critical contour for a wide class of rendering functions $\mathcal{C}$, we proved:

\begin{theorem}
Given a surface normal field $N(x, y)$ and any two choices of generic rendering functions $F_1, F_2 \in \mathcal{C}$, construct $I_1 = F_1(N(x, y)), I_2 = F_2(N(x, y))$.  If a critical contour is present in $I_1$, then there is a arbitrarily close critical contour in $I_2$.
\end{theorem}

\noindent
A corollary of this statement (Corollary 10 in \cite{Kunsberg18}) can be restated:

\begin{corollary}{}
An extremal contour must lie in the tubular neighborhood of a critical contour and have the same endpoints.
\end{corollary}

As extremal contours lie near the critical contours, the invariance statements from \cite{Kunsberg18} show that extremal contours will nearly always be salient from the image, regardless of rendering function.  Extremal contours delineate the surface features while the critical contours are salient from image.  We can observe a critical contour, infer an extremal contour next to it, and then use the previous section to attribute surface protrusions to image regions.





\begin{figure}
\begin{center}
\includegraphics[width=15cm]{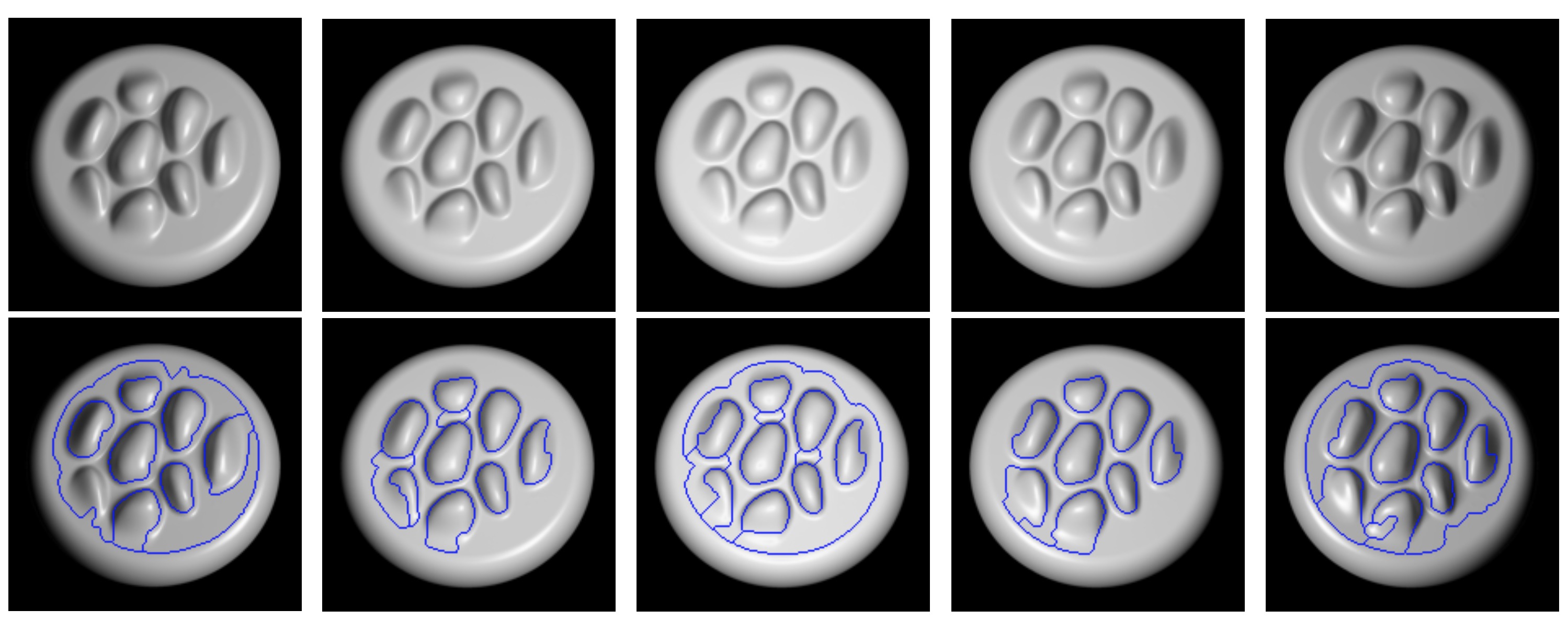}
\caption{Critical contours on differently rendered images by persistence simplification. Top Row:  A sequence of images of a surface rendered with specular and Lambertian shading;  the light source moves from left to right.  Bottom Row:  Closed critical contours on the image; our theory implies that these will contain 'bumps'.  Note that some bumps are "connected" by 1-cells. Although these are part of the simplified MS-complex, they are not critical contours. \label{fig:cobblestones}}
\end{center}
\end{figure}

Thus far we have only considered single bumps. But since the MS-complex is global, multiple bumps can be identified from portions of the MS-complex. We provide an example of this in Fig.~\ref{fig:cobblestones}. This is elaborated upon in the next Section, where we discuss computational issues.

\section{Computing Extremal Contours}

There are different ways in which our theory of extremal contours can be realized. We first show results from what is the standard computational approach. This was used in \cite{Kunsberg18} and to compute all of the examples in this paper thus far. We then develop what can be described as a more biological approach, which (perhaps surprisingly) involves concentric flows, and discuss some of its implications.

\subsection{Persistence Simplification}

Topology is the study of shape constancy under 'rubber sheet' distortions; i.e., the study of which properties (e.g., the number of holes) remain invariant under smooth deformations. As such, there is a concentration on issues such as when two shapes are (topologically) equivalent, e.g., when they have the same number of components and holes. Given the importance of geometric matters to this paper, two shapes are equivalent when their Morse-Smale complexes are equivalent. It is important to note that all of this is defined in the domain of continuous mathematics. But when noise and sampling are introduced, they introduce technical problems in assessing topological signatures: since pixels are discrete, what is the precise location of a singularity; since changes can only occur in the x- and y-directions (image coordinates), how can the actual gradient be specified; is this a real tiny hole in the shape or noise? The field of computational topology is being developed to answer them \cite{edelsbrunner2010computational, carlsson2009topology}. We highlight one of the important developments in this field  -- {\em persistence simplification} -- because this is the basis for the algorithm used in computing the examples in this paper.

Computational topology takes a discrete structure, covers it with a smooth object, and then uses it to calculate topological features. Just as blurring can "smooth over"  tiny holes due to noise in an image, persistence simplification is 
a globally consistent way to reveal overall structure while removing those tiny holes and 'irrelevant' noisy details that derive from quantization and discretization. In effect, it is a type of global, structure-preserving smoothing. Algorithms to compute the Morse-Smale complex from discretized images (i.e., a mesh) have been developed by
\cite{Reininghaus11, Sahner08, Weinkauf09, Weinkauf10}, among others. We use the algorithm of \cite{Reininghaus11} in all of our experiments.

We now show an example of our theory applied to a set of rendered images. Our goal is to show that, with the above ideas, one can segment the bumps and valleys of a surface without considering the rendering function.  We perceive accurately the bumps because we find critical contours in the orientation flow and associate them with extremal contours; in the special case when those extremal contours are closed, extremal rings provide the bump localization.

Fig.~\ref{fig:cobblestones}, we show images of a `cobblestone' (\cite{ToddVSS17})  rendered with both specular and Lambertian shadings; the light source is varied from left to right.  Each image has the same occluding contour.  The critical contours of the shading are computed for each image.  Note the tendency of these contours to surround the bumps. Persistence simplification is used to eliminate some artifacts (noise/discretization loops), but it is incomplete. The lateral condition of "steep sides" in intensity required for critical contours is not implemented, so sometimes issues of highlights, etc., can effect the result. Nevertheless, by and large almost all bumps are localized in every example. 

The next approach to computing extremal contours is more biological in its foundation. 

\subsection{Critical Contours and Circular Flows}

\begin{figure}
    \centering
    \includegraphics[width= 0.8 \linewidth]{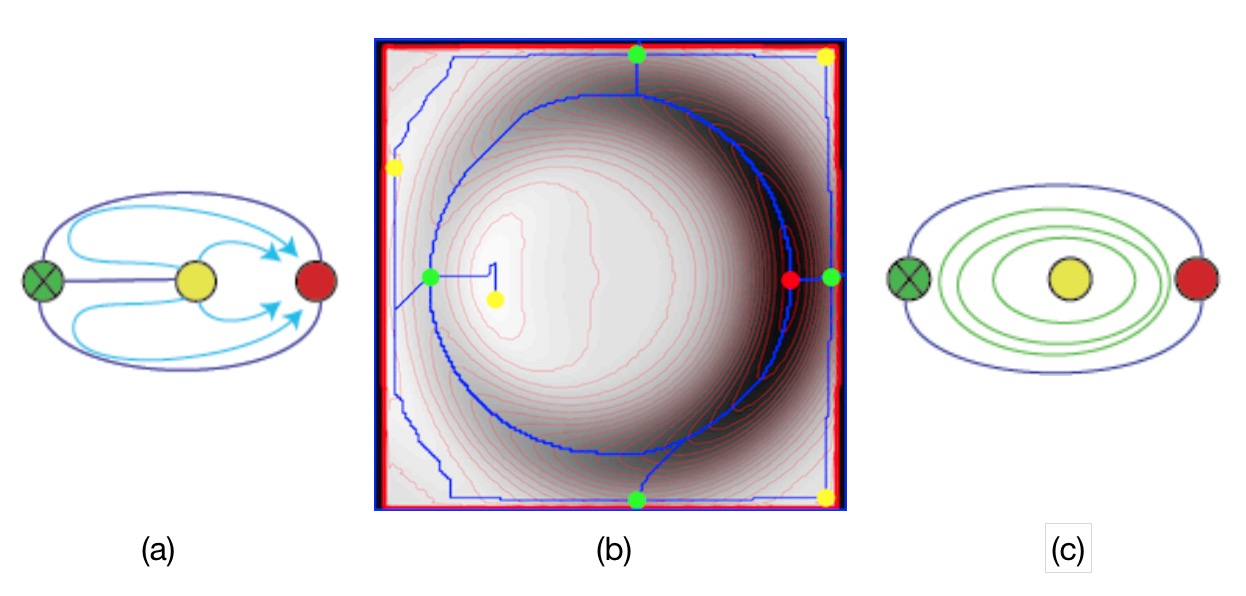}
    \caption{Concentration of isophote tangent orientation around bumps. (a) The slant extremal contour from Fig.~\ref{fig:moving-isophotes}. Note the gradient integral curves in cyan. (b) A corresponding image from Fig.~\ref{fig:sigmoid}. Notice how the closed critical contour (corresponding to the slant extremal contour) cuts across the isophotes. Moving slightly interior, the isophotes form concentric rings; this is cartooned in (c) by the concentric iso-slant contours in green.  Such families of nested level sets can serve as an image (or slant) signature for the identification of bumps.}
    \label{fig:nested-isophotes}
\end{figure}

A more biological approach to computing extremal contours would work directly on the orientation flow, rather than the discrete image. We illustrated earlier how isophotes and textures concentrate near the occluding contour (Fig.~\ref{fig:occluding}); similiar properties hold around bumps. In particular, focusing on the isophotes during the evolution shown in Fig.~\ref{fig:saddle-merge} makes clear that they concentrate in an analogous fashion near the critical contour; see Fig.~\ref{fig:nested-isophotes}. We are especially interested in those that form a concentric flow within the extremal ring.

Perceptual sensitivity to closed contours \cite{kovacs1993closed, elder1993effect} and circular textures (e.g., \cite{wilson1998detection, dumoulin2007cortical, dakin2002summation} is well known.  Importantly, as with these concentric flows it is not particularly sensitive to positioning and other perturbations (e.g, \cite{achtman2003sensitivity}). Furthermore, there is an accumulating body of evidence that orientation structure is at the basis of related shape perception \cite{li2000perception, fleming2011estimation, marlow2019photogeometric}, and is interpreted as a component of shape representation at an intermediate stage \cite{gallant2000human}. This has a particular advantage from our perspective, because our result formally connects an image salient property with a three-dimensional shape interpretation. This clearly opens the door for intermediate visual areas such as V4 to be sensitive to structure in 3D as well. 

\begin{figure}
\begin{center}
a) \includegraphics[width= 0.8 \linewidth]{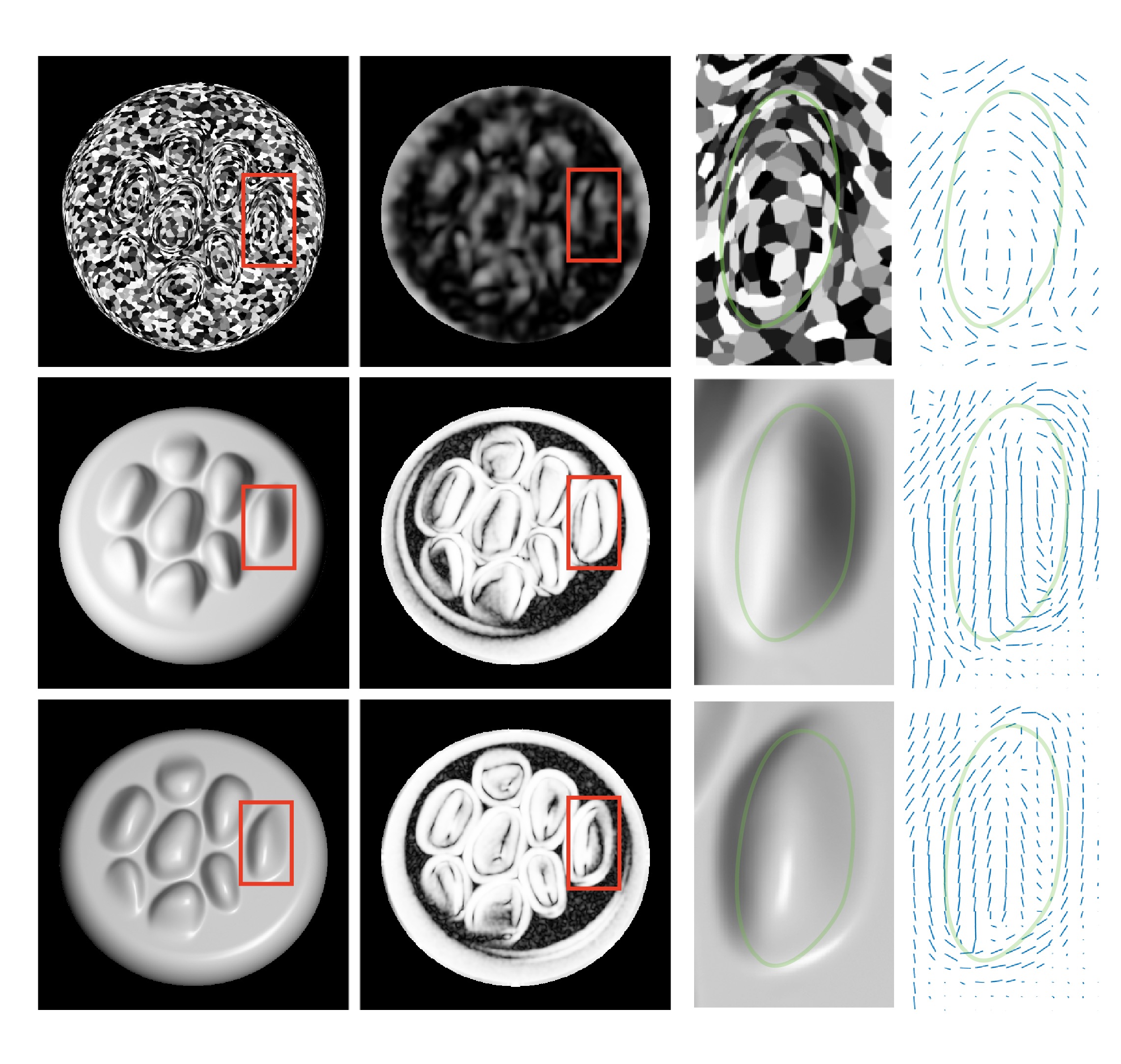}
\caption{Max slant cycles can be found via maximum compression in shading and textured images. Across each of the rows, we show: original image, 'flow compression', an enlarged region and the 'flow direction'.  Notice the similarities in the 'flow compression' and 'flow direction' across the different cues. 
 \label{fig:texture}}
\end{center}
\end{figure}

\begin{figure}
\begin{center}
\includegraphics[width= 0.8 \linewidth]{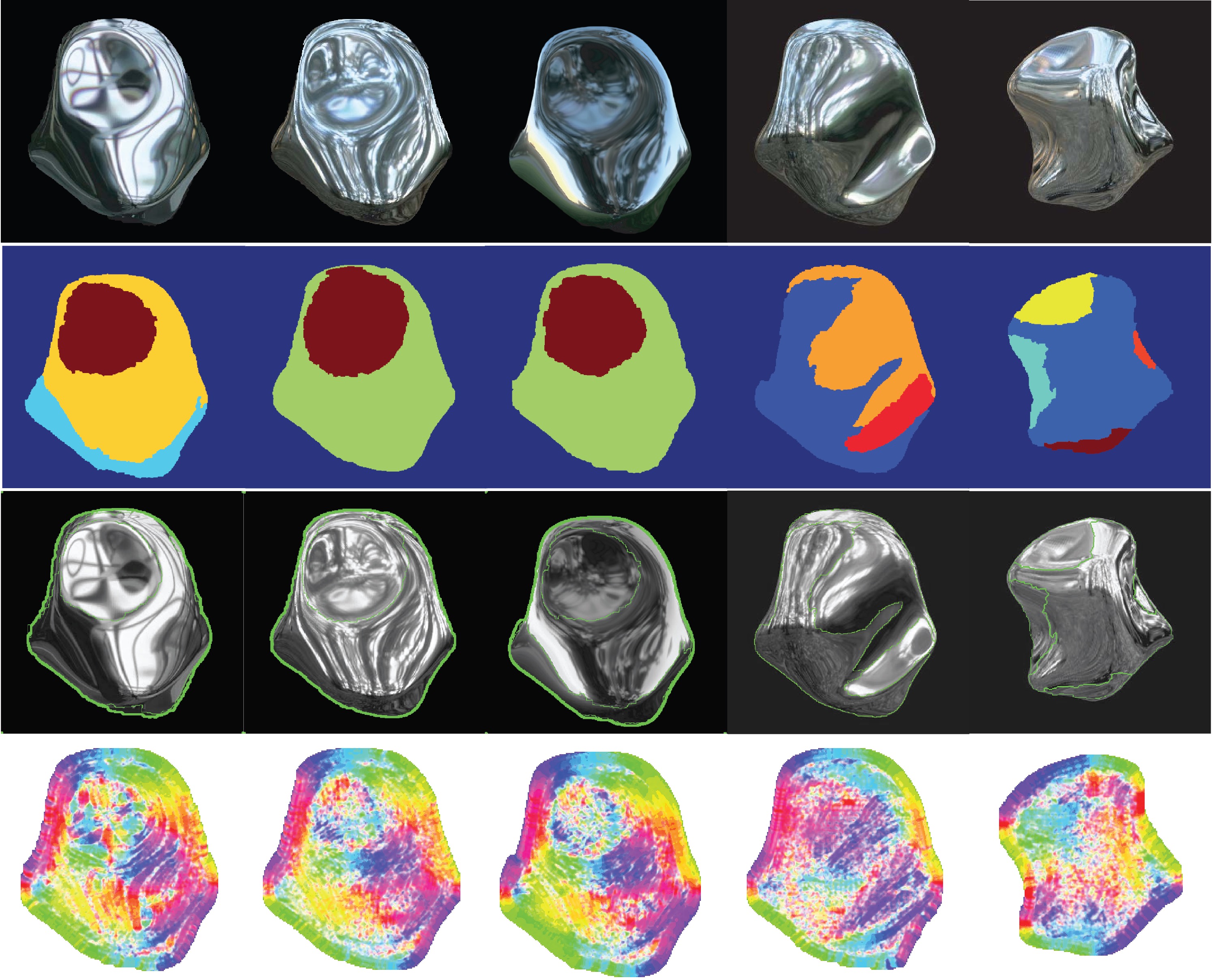}
\caption{Extremal curves on specular images (from \cite{fleming2004specular}).
(Row 1) Examples of five specular images. Note the organized flows apparent around the bumps and dents.
(Row 2) A segmentation of the images according to integral curves through the structure tensor direction.
(Row 3) The integral curves.
(Row 4) The colorized orientation map.
\label{fig:specular}}
\end{center}
\end{figure}

In Fig.~\ref{fig:texture} we show the oriented shading flow around two pebble examples, plus an integral curve\footnote{An integral curve through a vector field is a curve whose tangent agrees with the vector at each point.} through it. Notice how this curve surrounds the bumps just as the extremal ring did. Working with nested flows also supports a generalization from shape-from-shading to shape-from texture, because the texture compression works in a similiar fashion to produce nested flows around bumps (Fig.~\ref{fig:texture}). Finally, we show a number of specular examples, in which the oriented flows arise from compression of the visual scene around the object \cite{fleming2004specular, Mooney14}; see Fig.~\ref{fig:specular}.

In all of these examples, the flow compression was computed by a classical technique in image processing, the structure tensor \cite{rao1991computing, bigun1991multidimensional}, from which the primary direction was derived. Compression was assessed by the ratio of eigenvalues. In effect, the structural tensor is an ellipse, and the compression is a measure of elongation. There are more biological ways to compute these flows \cite{ben2004geometrical}, but the structure tensor serves as an easy proof-of-concept. 

\section{Implications and Discussion}

We now illustrate some uses of the above theory proposing that extremal contours are invariant, salient, and a useful abstraction to understand the surface.  We have just illustrated how the theory can be used to find `bumps' in images with unknown rendering and illumination.  Now, we show how the theory predicts a novel 3D bistable illusion. Following, we show an illustrative example applying the theory to non-closed extremal curves; a constraint labeling problem arises. Finally, we show a demonstration comparing the importance of these extremal contours over other image regions.

\subsection{Bi-Stable Dimples and Bumps}

The definitions of extremal contours and rings were in terms of the MS-complex, and involved only singularities and their (global) relationships. There is often a natural relationship between these critical points in the slant field and image field. For example, maxima in the image can correspond to minima in slant, which follows from Lambertian (and other) rendering functions.





\begin{figure}
\begin{center}
\includegraphics[width=1 \linewidth]{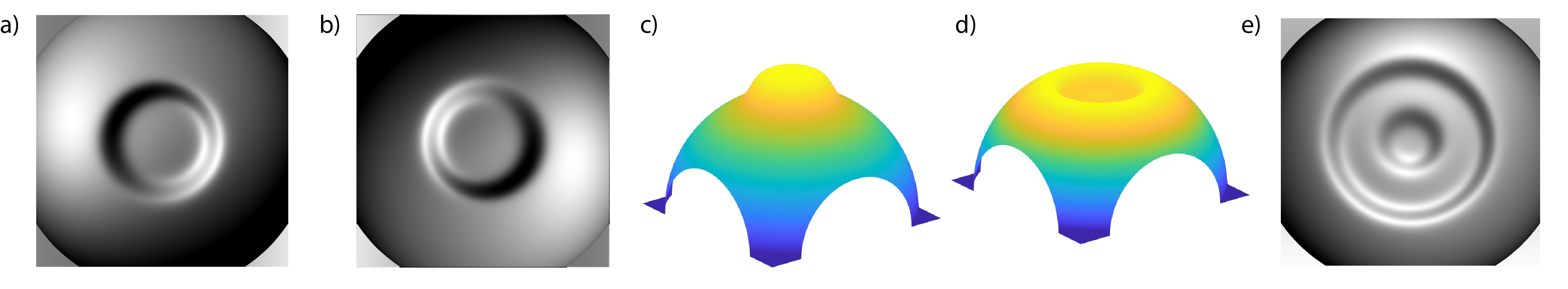}
\caption{\label{fig:bistable} Multi-stable bumps and dents. 
(a, b): Images with bistable perceptions governed by an extremal curve segmentation. The inner disk is carved out by an extremal contour and can be seen as coming out or going into the page. At first glance, (a) often looks like an indentation and (b) like a protrusion; as in the crater illustion, (b) is the inversion of (a). Staring at either one shows that it can flip to the other. (c, d): Possible surface interpretations of images in (a and b). (e): Nested bumps provide a higher-order multi-stable illusion. Again, components can be flipped independently.}
\end{center}
\end{figure}

\begin{figure}
\begin{center}
\includegraphics[width=0.7 \linewidth]{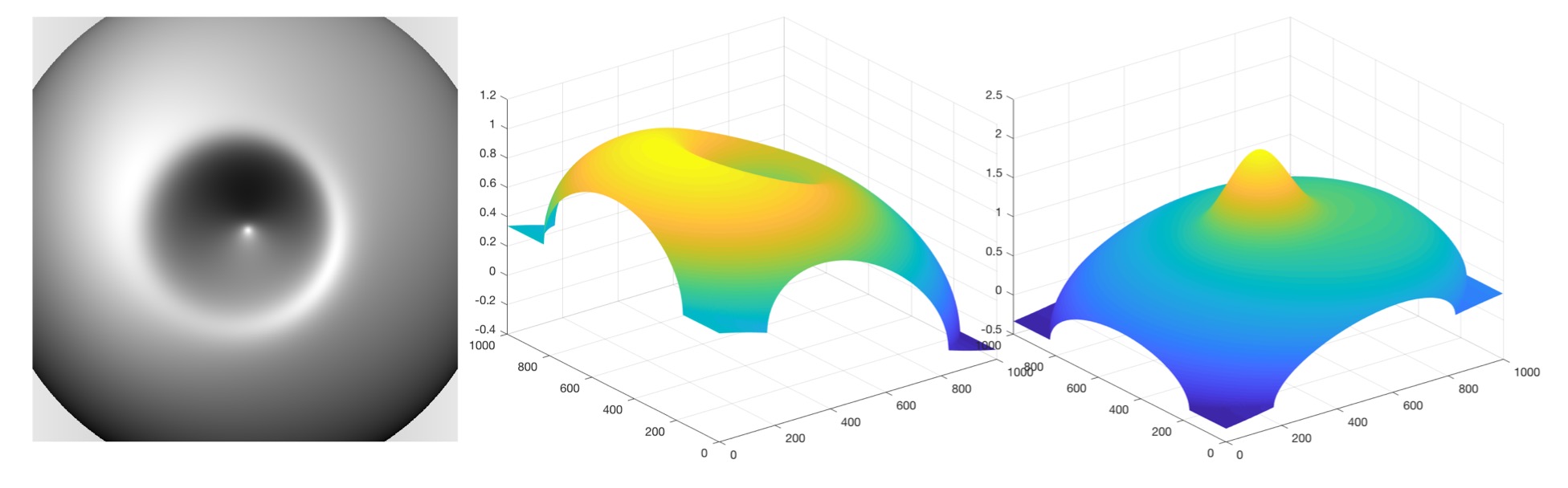}
\caption{\label{fig:bistable2} An extension of Fig. \ref{fig:spherical_ridge}, in which the bump is not directed at the
viewer and the object is asymmetric. Bi-stability still maintains.}
\end{center}
\end{figure}

The existence of bumps and valleys, as interpretations of extremal contours, immediately brings the convex/concave illusion to mind \cite{ramachandran1988perception}.
Set up properly, this is a classical instability in shape perception, often disambiguated by the common light-source-from above heuristic.  
Since an extremal ring perceptually creates either a bump or a valley, it follows that similar instabilities should be demonstrable here as well, which further highlights the inherent ambiguity.

But unlike changing the global light source, an important difference arises: since the extremal contour actually segments the bump or valley, these segmented parts should be independent of other portions of the image. This raises a prediction: that the individual parts of an image should also be subject to the multi-stability {\em individually}. 

To confirm this prediction, i.e. to show that the individual parts -- and not the full image -- can be flipped separately, we created the bi-stable images in Fig.~\ref{fig:bistable} with an extremal ring in the center whose interior region is a disk. The extremal ring creates a perceived segmentation, but, as described in Section \ref{sec:ELSM}, does not specify whether the interior disk is a bump or a valley. The normal vectors along the ring (once projected into the image) can point into or outside the ring. Thus, the interior part can be seen as a bump (pointing out of the page) or a valley (into the page) and can be perceptually flipped.  An asymmetric version of this illusion is shown in Fig.~\ref{fig:bistable2}. We see that a segmentation is necessary for describing this perceptual phenomenon; it is cleaner than representing the two solutions as independent depth fields. Significantly, the bi-stable bumps can be nested as well.

\begin{remark}
This is not the concave/convex (`hollow face') illusion in disguise because the surface portion outside the extremal ring is stably perceived as convex in depth.  This is due to the portions of the occluding contour that are visible at the edges of the image.  This example could be generalized to include $N$ extremal `rings' leading to a possible $2^N$ ambiguous perceptions.  Rather than a single concave/convex ambiguity on the global object, we could have concave/convex ambiguity on individual parts governed by the extremal curves.
\end{remark}

\subsection{Interpreting Extended Extremal Contours}

\begin{figure}
\begin{tabular}{c c}
\includegraphics[width=0.4 \linewidth]{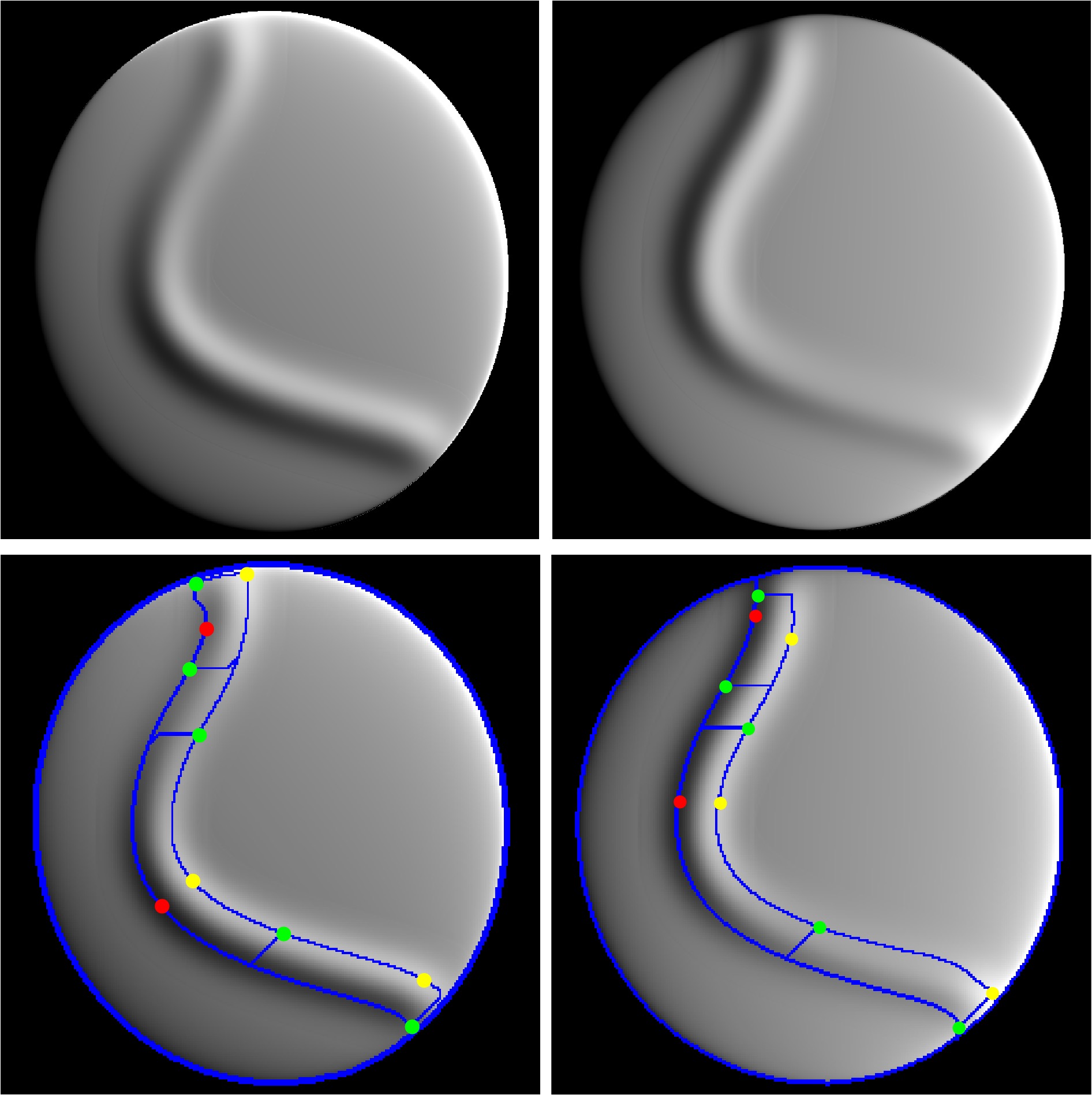} &
\includegraphics[width=0.4 \linewidth]{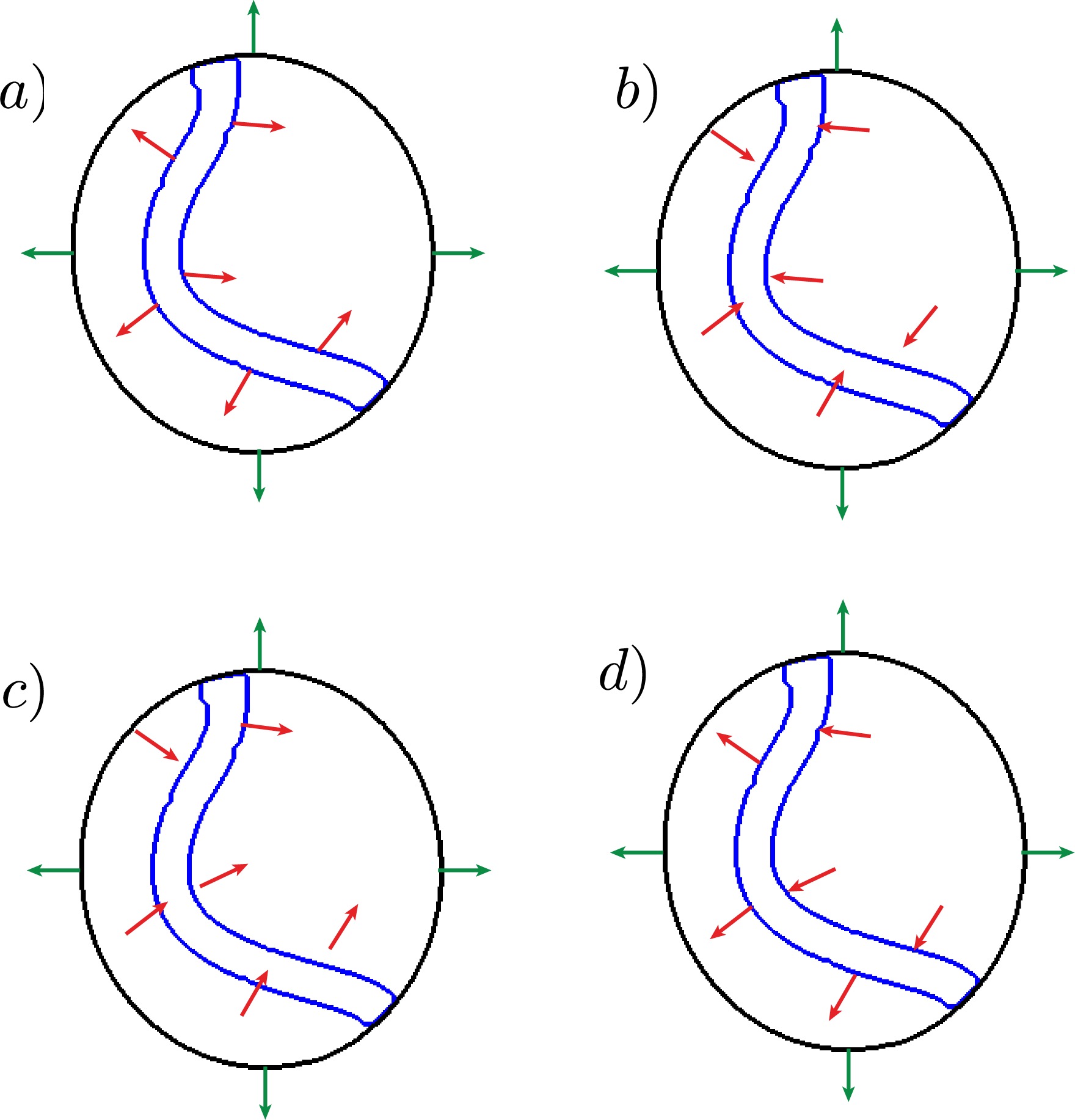} \\
(A) & (B) \\
\includegraphics[width=0.45 \linewidth]{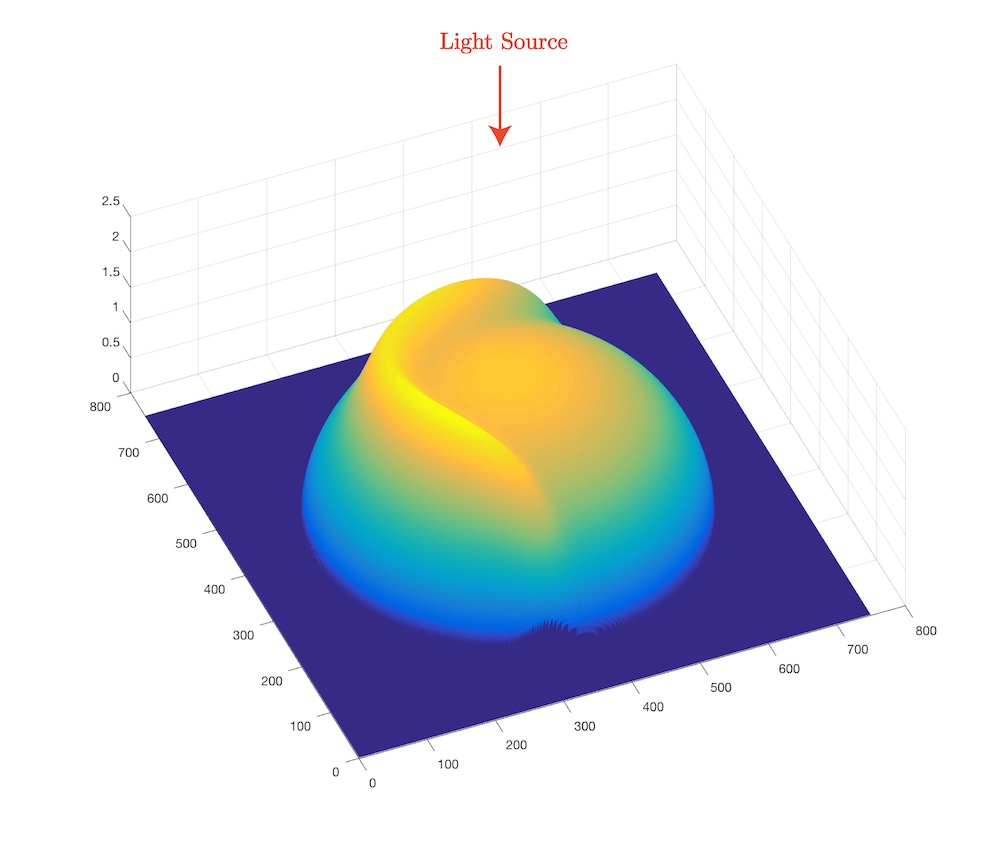} &
\includegraphics[width=0.45 \linewidth]{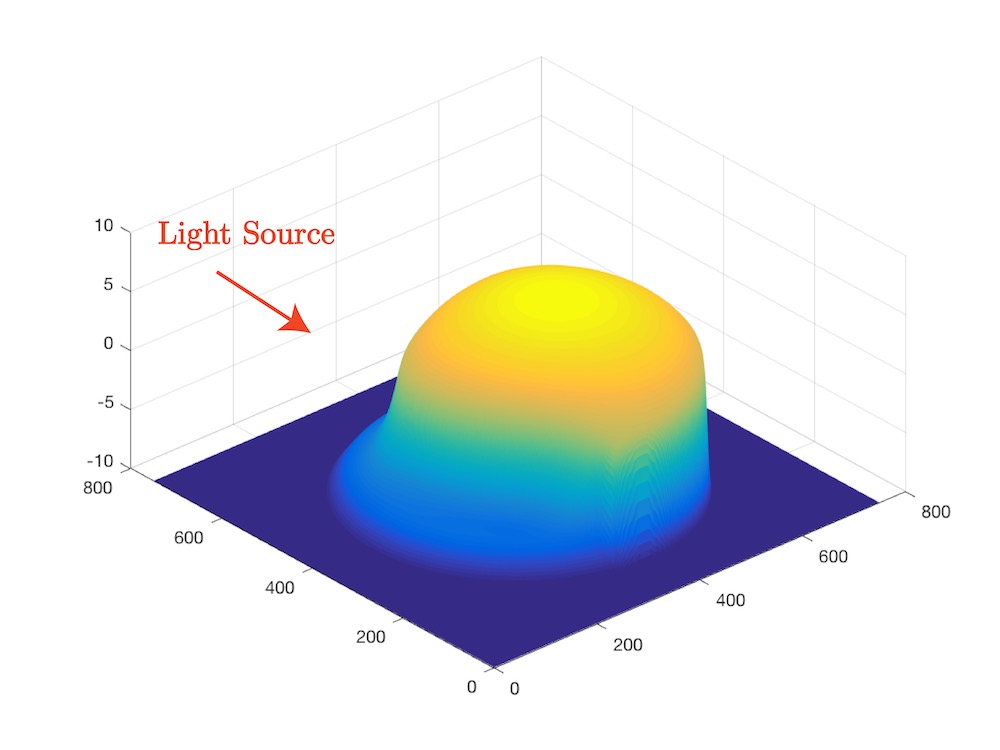} \\
(C) & (D) 
\end{tabular}
\caption{ From bumps to extended ridges. (A) Top Row: Images of the same surface illuminated from different positions.  Bottom Row: Persistence-simplified Morse-Smale complexes showing the extremal curves in blue.  (The short horizontal segments should be ignored; they are not extremal contours.) (B) Four possible labelings of the surface normal field on these curves. (Ba) A ridge above the surface; (Bb) a valley; (Bc, Bd) two extended slopes. (C) The surface corresponding to (Ba). (D) The surface corresponding to (Bd). \label{fig:spherical_ridge}}
\end{figure}

The bistable illusion above arises from the binary choice of the normal field on the extremal ring -- it can either point consistently toward the "inside" (as in a dent) or the "outside" of the bump. We previously argued that, to be generic, such consistency is required almost everywhere along an extremal contour.  We now elongate the bumps to show how a combinatorial calculus of normals develops. 

Individual non-closed slant extremal curves can still provide information about protrusions and other features within a shape.  The result in Section~\ref{sec:ELSM} can be summarized as: the normal field along an extremal curve must point to one side of the curve (almost always). This can be thought of as a labeling of the curve, just as the "border" side of a curve is indicated by the Gestalt notion of "border ownership" \cite{zhou2000coding}. If extremal curves are extended toward an occluding contour without introducing other structure, then the logic can be applied. For each of the two extremal curves in the example surface shown in Figure \ref{fig:spherical_ridge}A, we get the four qualitatively different labelings depicted in Figure \ref{fig:spherical_ridge}B.  The red arrows depict the binary choice for the surface normal on the curve and the green arrows depict the necessary normal constraints due to the occluding contour. These labelings are simply a choice of orientation for the surface normal on each of the two blue curves.  

\begin{figure}[]
\begin{center}
\includegraphics[width=0.8 \linewidth]{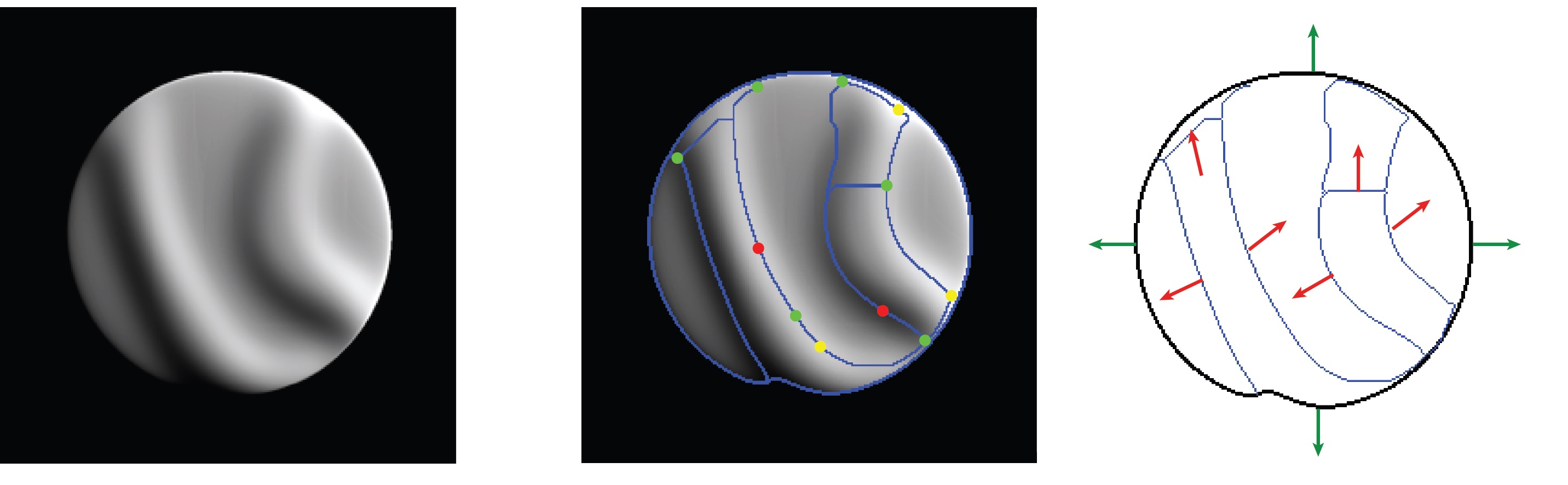}
\caption{\label{fig:spherical_ridge2} An extension of Fig. \ref{fig:spherical_ridge} in which the critical contours partition the surface into several parts. These can then be interpreted as ridges and valleys through a consistent labeling scheme.}
\end{center}
\end{figure}

Although we have not done formal psychophysics on this, the authors perceive labeling (a) although one can construct surfaces with precisely the same images in Fig.~\ref{fig:spherical_ridge}(A) from the other labelings also; see Fig.~\ref{fig:spherical_ridge}(D). Similiar labelings can be applied to more complex imagery (Fig.~\ref{fig:spherical_ridge2}). Further examination of these issues is warranted, at least to extend our understanding of generic lighting and surface interactions.

\subsection{Critical Contours as 3D Shape Anchors}

Our basic hypothesis is that critical contours are key to 3D shape perception. Thus far we have shown theoretically that they provide a scaffold on which qualitative shape inferences can be based, and that the segmentation they provide can lead to multi-stable shape displays. In a related paper \cite{Kunsberg2018Focus} we exploited a color-shape interaction to demonstrate that the neighborhood around critical contours, and not the space between them, sufficed to ground 3D inferences. We now provide additional support for this observation with two new displays.

\begin{figure}[h!]
\centering
\includegraphics[width=15cm]{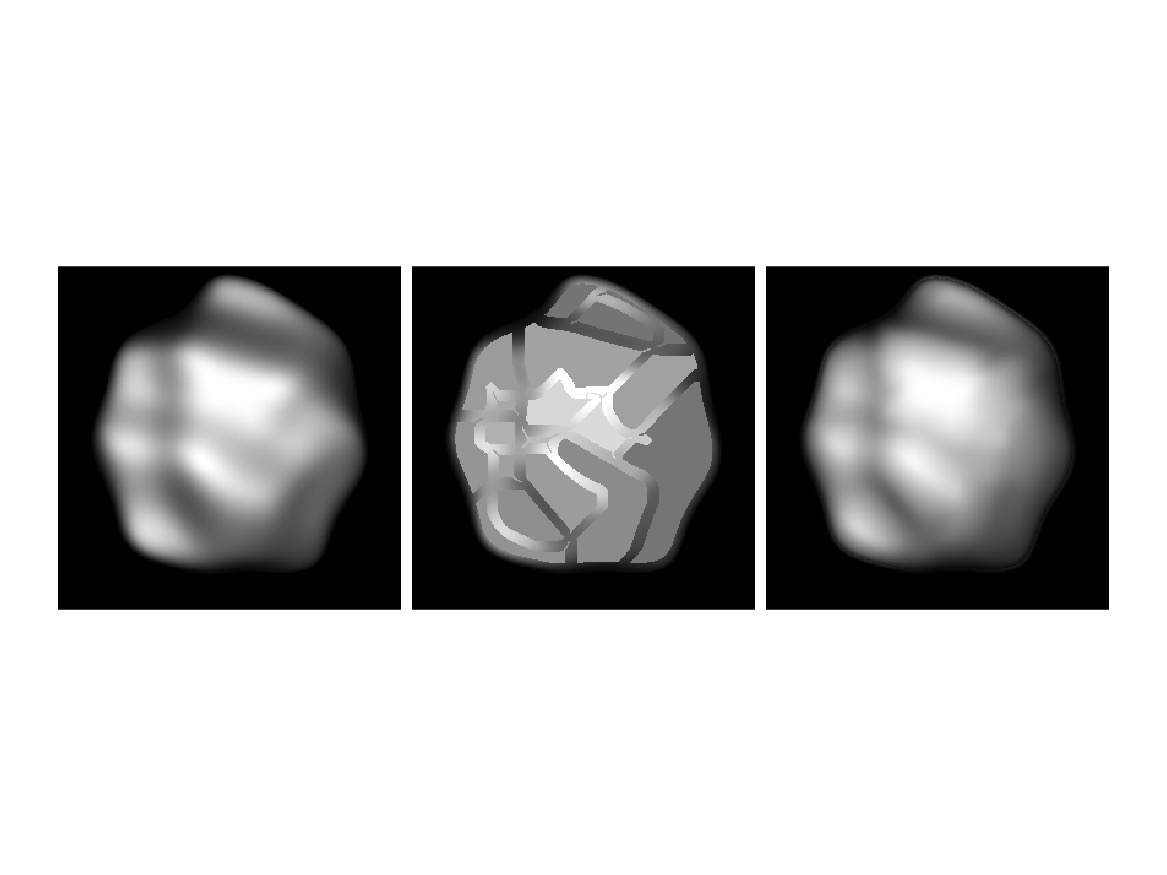} \newline
\includegraphics[width=8cm, trim={6cm 0cm 6cm 3cm},clip]{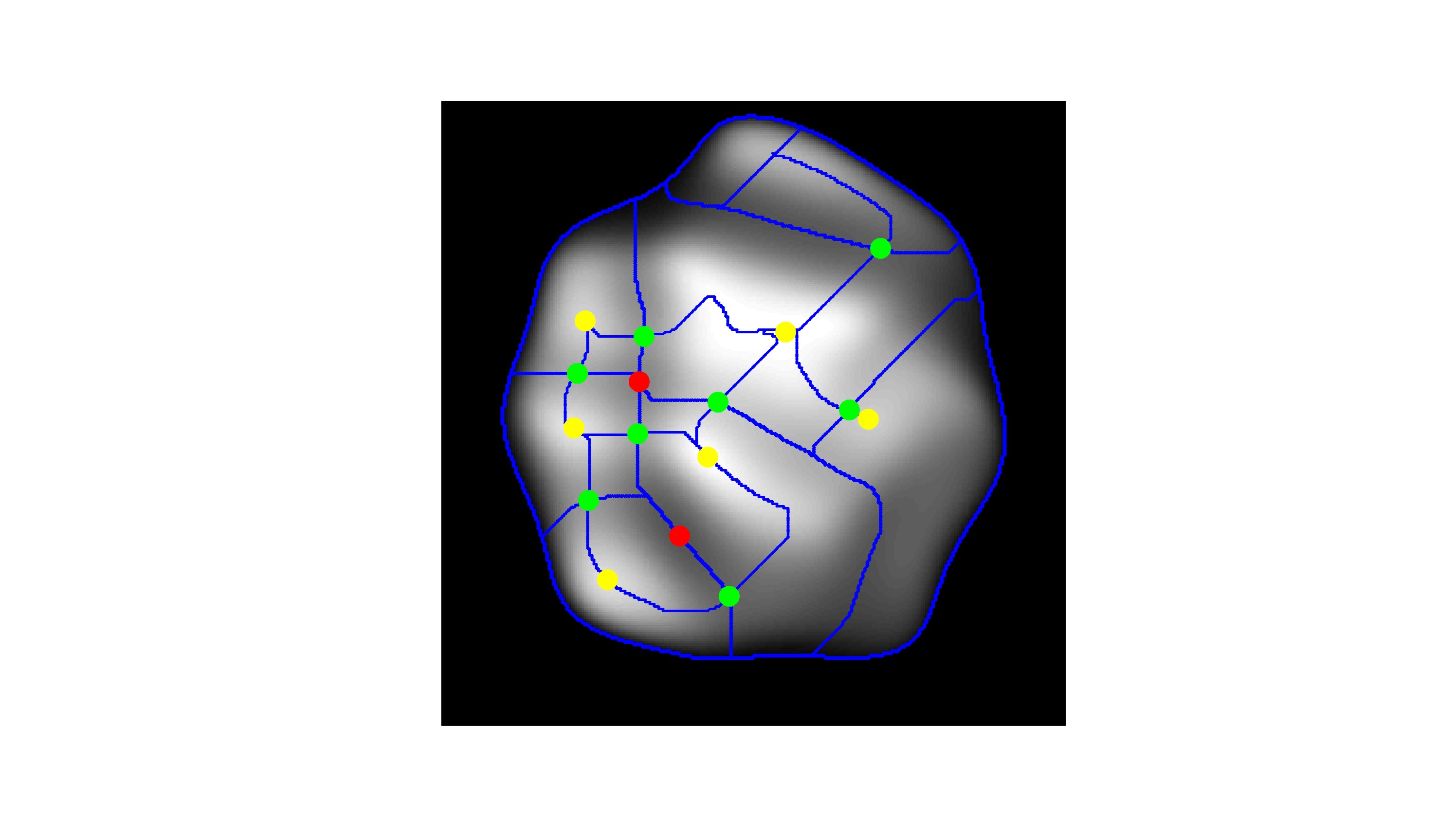} 
\includegraphics[width=8cm,  trim={2cm 0cm 0cm  3cm}]{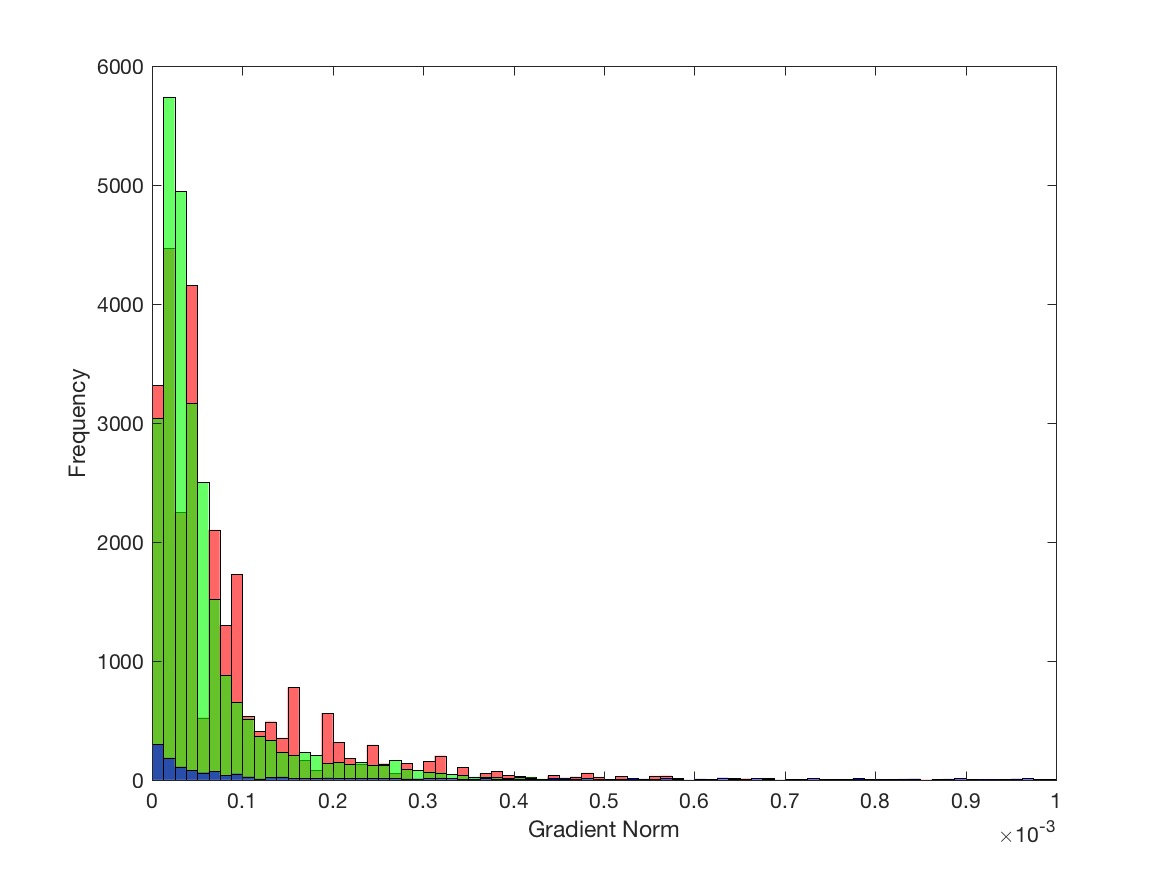} 
\caption{\label{fig:cc_importance1} (top row) (left) A random blob with a vivid shape percept.  (middle) A flattened image. The intensities on the MS complex match the image intensities while, for each region, the intensity is replaced by the average. (right) A pseudo-heat equation blur of the middle image.  (lower row) (left) The MS complex. (right) Histogram of intensities comparing the images in the top row. Red corresponds to left;  Blue corresponds to middle and Green corresponds to right.  }
\end{figure}

The placement of critical contours is key to 3D shape perception.  We first illustrate the relative unimportance of shading values away from the critical contours. We first isolate the critical contours, and then set the intensity along them to their underlying value. The areas between critical contours (the 2-cells) are set to a constant, which is the mean value of intensity within it. Finally, the resultant discontinuous image is smoothed by a pseudo heat equation. While the intensities in the resultant image differ from those in the original by a non-linear function, the shapes appear about the same; see Fig.~\ref{fig:cc_importance1}. 

\begin{figure}[]
\begin{center}
\includegraphics[width=0.2 \linewidth]{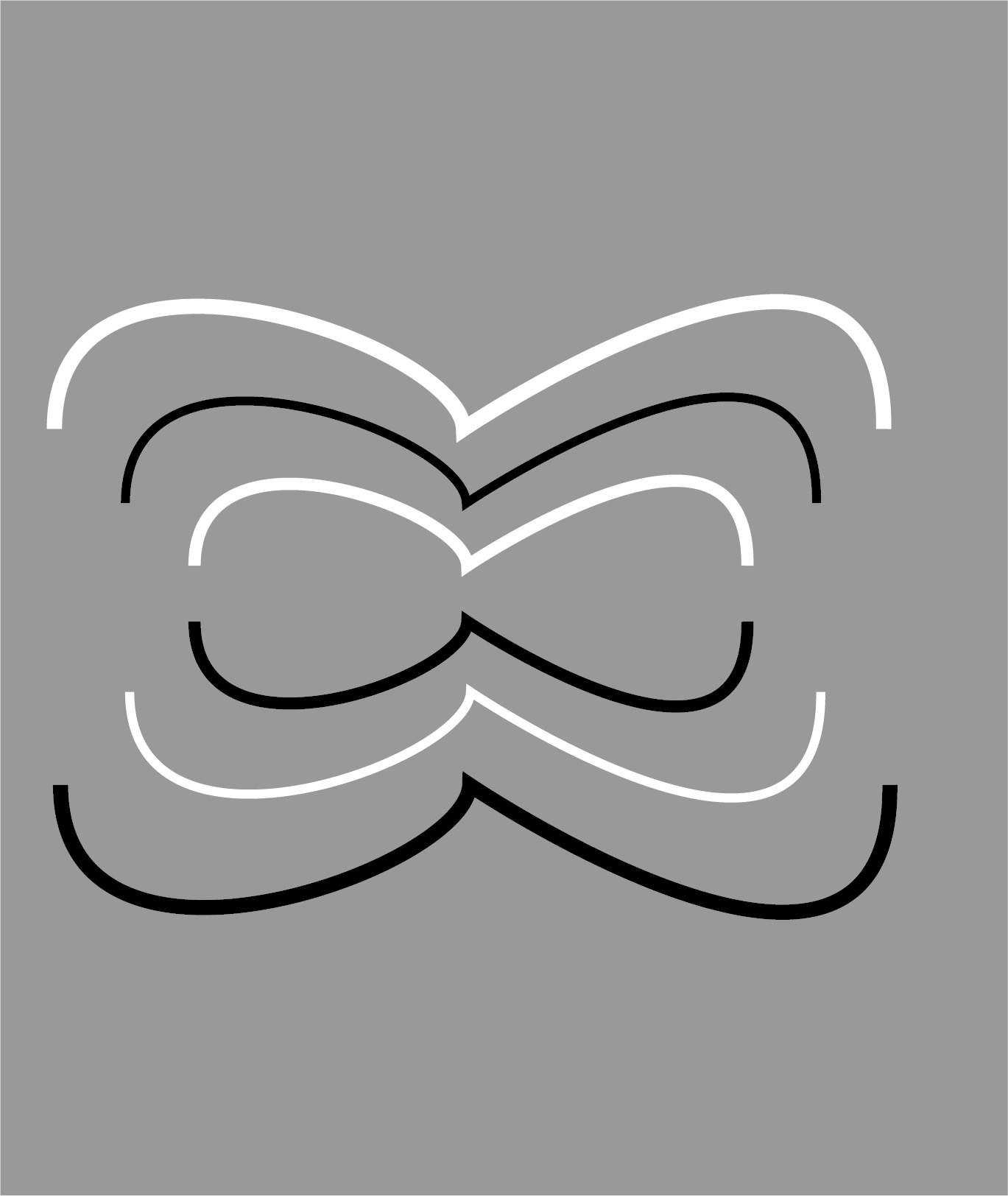}
\includegraphics[width=0.4 \linewidth]{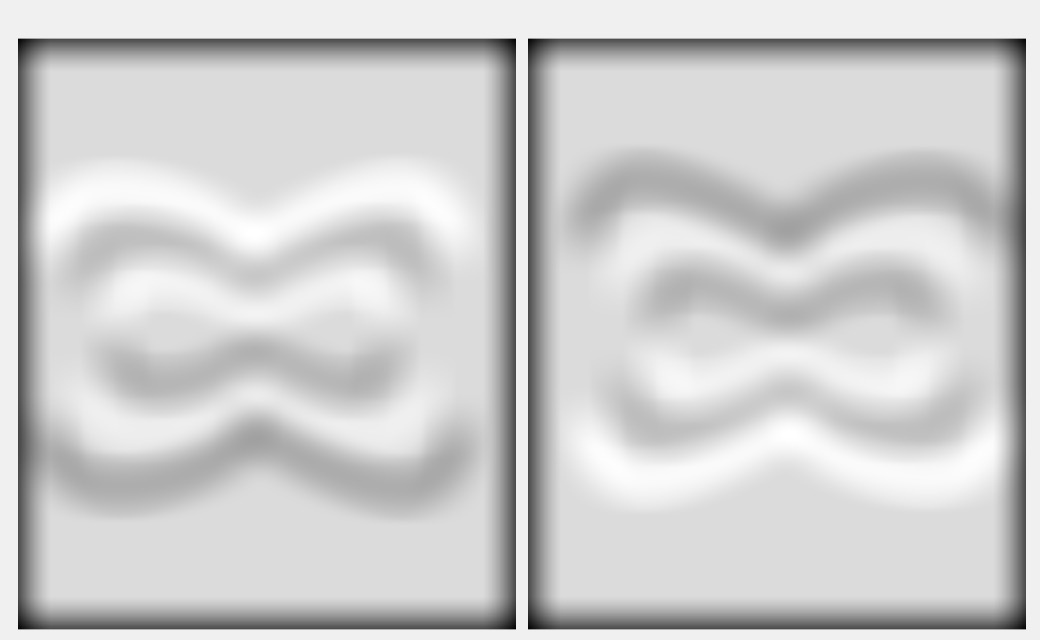} \\
\caption{\label{fig:cc_importance2} Left: A set of image contours in black and white on a grey background.  Center: A simple blurring operation on these contours yields a 3D percept.  Note the perception of ridges and bumps delineated by the orginal contours.  Right:  If the original contours are inverted (black to white and vice versa), the percept is also inverted.}
\end{center}
\end{figure}

The second demonstration in Fig.~\ref{fig:cc_importance2} complements the above. We begin with a set of contours of constant intensity against a neutral background. Blurring these provides a rich 3D percept with the same multi-stability properties discussed previously. 

While these two demonstrations are only suggestive, we include them to (hopefully) stimulate others to examine these questions more rigorously.

Conversely, we can construct a 3D percept by solely drawing dark and bright curves on a grey background and blurring.  The 3D percept contains ridges, bumps, and valleys exactly where predicted by our theory.

\section{Conclusions}

The visual system effortlessly computes qualitative 3D shape without knowing the rendering function. Our solutions are robust across many lighting and material compositions. We have argued that, to understand such remarkable phenomena, the right qualitative 3D surface representation is required, and it must be linked to measurable 2D image features that remain invariant.  In this paper, we advocated for the use of slant extremal curves as the candidate surface representation coupled with critical contours providing the invariant 2D features. 

Image structure is highly dependent on the object material and rendering function.  For example, a simple change in illumination will cause the vast majority of image pixels to change.  In addition, rendering function changes (e.g. Lambertian vs. specular) can lead to unpredictable pixel changes in most of the image. If the visual system's 3D shape inferences equally weighted every pixel, then a uniformity arises. Examples of this include the \emph{inverse optics} approaches (e.g. \cite{horn89}), which are notoriously unstable.  Rather, by focusing on extremal slant curves, we ignore many of the `nuisance' parameters in the image associated with specific rendering functions.  In the process the computation becomes more straightforward and biological:  there is a direct route from local image orientations to global surface parts without having to e.g. solve a particular differential equation.

Much work remains to be done. First, the theory of critical contours and extremal curves leads to psychophysical tests; we mention the predictions inherent in the multi-stable displays and the extended ridges. Initial demonstrations show that those regions away from extremal curves do not carry much 3D information (Figs.~\ref{fig:cc_importance1}, ~\ref{fig:cc_importance2} and \cite{Kunsberg2018Focus}). Second, computational issues remain. The image extremal contours may be incomplete due to noise and occlusion; some type of contour completion may be needed. Also, as shown in Fig.~\ref{fig:spherical_ridge}, a labeling algorithm may be needed to compute the `side' to which the normal field points.  
Finally, the transition from the critical contour "scaffold" to a full surface realization needs to be studied. Initial experiments on reconstruction of the entire slant field from its extremal curves are promising  \cite{Kunsberg18}, and related work exists in the computational literature (e.g., \cite{Giorgis15, Weinkauf10}). Whether this is analagous to the filling-in processes around Kanizsa figures, neon color spreading, and so on \cite{van2009filling, pessoa2003filling, bressan1997neon} remains an enticing possibility. 

In summary, we have shown a path toward 3D shape inference. It captures the qualitative nature of shape perception while guaranteeing a degree of constancy even when the rendering function is unknown.

\newpage

`

\section{Appendix: Supplementary Material}
\subsection{Review of the Morse Smale Complex}
\label{M-S}
The Morse Smale complex is a qualitative representation emphasizing the different stable and unstable regions of a smooth scalar function.  In this work, we choose the function to be the slant function of the image surface $\sigma(x, y): \mathbb{R}^2 \rightarrow \mathbb{R}$.  We will assume $\sigma$ is a \emph{Morse function}: all its critical points are non-degenerate (meaning the Hessian at those points is non-singular) and no two critical points have the same function value.  

For a smooth surface, the \emph{gradient} 
$\nabla \sigma = \left( \partial f / \partial x, \partial f / \partial y \right)$
exists at every point.  A point $p \in \mathbb{R}^2$ is called a \emph{critical point} when $\nabla \sigma (p) = 0$.  This gradient field gives a direction at every point in the image, except for the critical points, a set of measure zero. Following the vector field will trace out an \emph{integral line}. These integral lines must end at critical points, where the gradient direction is undefined.  Thus, one can define an \emph{origin} and \emph{destination} critical point for each integral line.  

The type of each critical point is defined by its \emph{index}: the number of negative eigenvalues of the Hessian at that point.  For scalar functions on $\mathbb{R}^2$, there are only three types: a maximum (with index 2), a minimum (with index 0) and a saddle point (with index 1).  

There are two types of integral lines, depending on the difference in index of the critical points it connects.  If the difference is one, we call the integral line a \emph{1-cell}.  It naturally must connect a saddle with either a maximum or a minimum.   For example, a \emph{saddle-maxima 1-cell} connects a saddle and a maximum.  The set of 1-cells will naturally segment the scalar field into different regions, called \emph{2-cells}.   In addition, the scalar values on the 1-cells govern the values on the 2-cells.  See Fig \ref{fig:guylassy} for insight.

Further, for each critical point,  its \emph{ascending manifold} is defined as the union of integral lines having that critical point as a common origin.  Similarly, its \emph{descending manifold} is the union of integral lines with that critical point as a common destination.   


 For two critical points $p$ and $q$, with the index of $p$ one greater than the index of $q$,  consider the intersection of the descending manifold of $p$ with the ascending manifold of $q$.  This intersection will be either a 1D manifold (a curve called a 1-cell or watershed) or the empty set.   For two critical points $r$ and $s$, with the index of $r$ two greater than the index of $s$, the intersection of the descending manifold of $r$ with the ascending manifold of $s$ will either be a 2D manifold (a region called a 2-cell) or the empty set.  Thus, the intersection of all ascending manifolds with all descending manifolds partition the manifold $\mathbb{M}$ into 2D regions surrounded by 1D curves with intersections at the critical points.

The Morse Smale complex is the combinatorial structure (and the corresponding attaching maps) defined by the critical points, 1-cells and 2-cells.  It is a structure that relates a set of contours (the 1-cells) to a qualitative function representation.  With knowledge only of the slant function at the critical points and 1-cells, one could reconstruct the 2-cells (and thus the entire function) relatively accurately.  For some insight, see \cite{Giorgis15, Weinkauf10}.  In this work, we wish to show how the slant saddle-maxima 1-cell can be used as a `bump boundaries' to model 3D shape perception.

For additional information, see \cite{milnor2016morse, Gyulassy08, Biasotti:2008:DSG:1391729.1391731, matsumoto2002introduction}.




\subsection{The Gauss Map}

\begin{figure}[]
\begin{center}
\includegraphics[width=0.5 \linewidth]{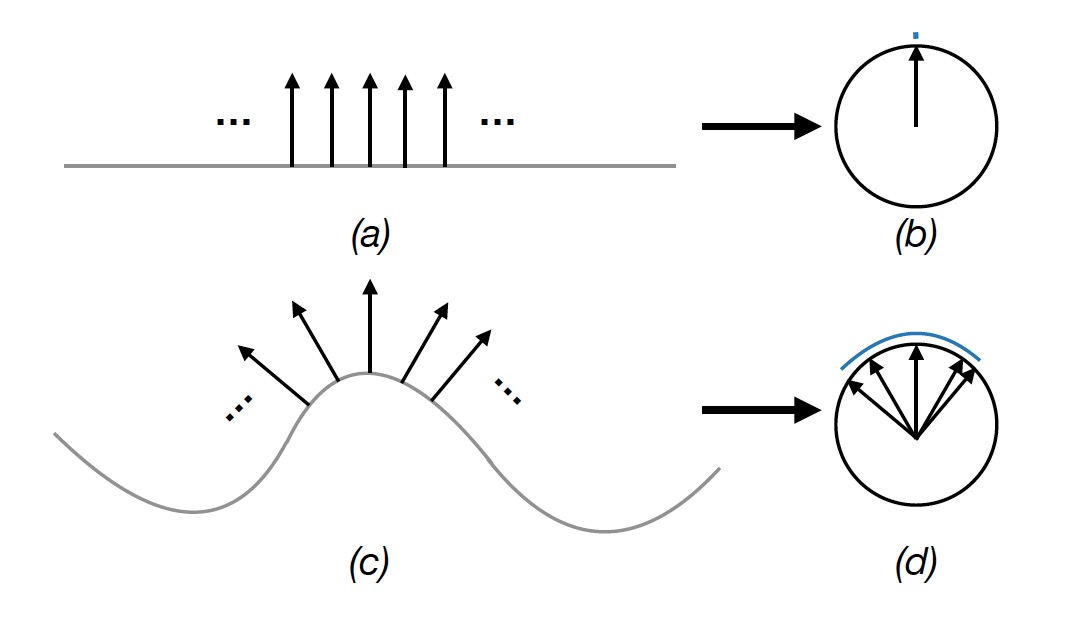}
\caption{Illustration of the Gauss map, which takes the normals attached to a curve (or surface) (left) and maps them to the unit circle (sphere) (right). (a) A selection of normals attached to a straight line; (b) these map to a single point on the Gauss circle. (c) A selection of normals attached to a curve; (d) these map to a region (in blue) on the Gauss circle. \label{fig:gauss-map-intro}}
\end{center}
\end{figure}

We use the Gauss map as an indication of how wildly a surface is varying. We now provide a brief introduction to it (Fig.~\ref{fig:gauss-map-intro}). For a more serious introduction, see \cite{oneill}.

Gauss, working on the foundations of curvature in differential geometry, designed a map that takes the normals to a curve or a surface and maps them, collectively, to a circle or a sphere. Intuitively, the map is accomplished by moving each of the (unit) normals to a single point. Notice how, for the (straight) line the normals then all overlap, while for the curve they "spread out" somewhat unevenly. This spreading out can be used as the foundation for a definition of curvature: for a given length of curve, the normals spread out over a portion of the Gauss circle; in the limit as this length of curve approaches zero, the area on the Gauss circle also approaches a limit. The ratio of these two areas is the curvature. Since this limit is taken around a point on the original curve, the curvature is a local descriptor. We exploit the measure of the normals over a region (of a surface) to get a global measure of variation.

\newpage

\printbibliography

@book{edelsbrunner2010computational,
  title={Computational topology: an introduction},
  author={Edelsbrunner, Herbert and Harer, John},
  year={2010},
  publisher={American Mathematical Soc.}
}

@article{lee1984two,
  title={Two-dimensional critical point configuration graphs},
  author={Nackman, Lee R},
  journal={IEEE Transactions on Pattern Analysis and Machine Intelligence},
  number={4},
  pages={442--450},
  year={1984},
  publisher={IEEE}
}

@article{smale1961gradient,
  title={On gradient dynamical systems},
  author={Smale, Stephen},
  journal={Annals of Mathematics},
  pages={199--206},
  year={1961},
  publisher={JSTOR}
}

@article{maxwell1870hills,
  title={L. on hills and dales: To the editors of the philosophical magazine and journal},
  author={Maxwell, J Clerk},
  year={1870},
  publisher={Taylor \& Francis}
}

@article{griffin1995superficial,
  title={Superficial and deep structure in linear diffusion scale space: Isophotes, critical points and separatrices},
  author={Griffin, Lewis D and Colchester, Alan CF},
  journal={Image and Vision Computing},
  volume={13},
  number={7},
  pages={543--557},
  year={1995},
  publisher={Elsevier}
}

@article{doi:10.1167/15.2.24,
author = {Egan, Eric J. L. and Todd, James T.},
title = {The effects of smooth occlusions and directions of illumination on the visual perception of 3-D shape from shading},
journal = {Journal of Vision},
volume = {15},
number = {2},
pages = {24},
year = {2015},
%URL = { + http://dx.doi.org/10.1167/15.2.24},
}

@article{doi:10.1068/i0645,
author = {James T. Todd and Eric J. L. Egan and Flip Phillips},
title = {Is the Perception of 3D Shape from Shading Based on Assumed Reflectance and Illumination?},
journal = {i-Perception},
volume = {5},
number = {6},
pages = {497-514},
year = {2014},

URL = { 
        http://dx.doi.org/10.1068/i0645}
}

@article{Seyama19983805,
title = "Shape from shading: estimation of reflectance map ",
journal = "Vision Research ",
volume = "38",
number = "23",
pages = "3805 - 3815",
year = "1998",
note = "",
author = "Jun'ichiro Seyama and Takao Sato",
}

@article{CHRISTOU19971441,
title = "Light Source Dependence in Shape from Shading ",
journal = "Vision Research ",
volume = "37",
number = "11",
pages = "1441 - 1449",
year = "1997",
issn = "0042-6989",
author = "Chris G. Christou and Jan J. Koenderink"
}

@article{Curran19961399,
title = "The effect of Illuminant position on perceived curvature ",
journal = "Vision Research ",
volume = "36",
number = "10",
pages = "1399 - 1410",
year = "1996",
note = "",
issn = "0042-6989",
author = "William Curran and Alan Johnston"
}

@article{doi:10.1068/p5807,
author = {Byung-Geun Khang and Jan J Koenderink and Astrid M L Kappers},
title = {Shape from Shading from Images Rendered with Various Surface Types and Light Fields},
journal = {Perception},
volume = {36},
number = {8},
pages = {1191-1213},
year = {2007},
URL = { 
        http://dx.doi.org/10.1068/p5807  },
}

@article{doi:10.1068/p251009,
author = {Jan J Koenderink and Andrea J van Doorn and Chris Christou and Joseph S Lappin},
title = {Perturbation Study of Shading in Pictures},
journal = {Perception},
volume = {25},
number = {9},
pages = {1009-1026},
year = {1996},
URL = { 
        http://dx.doi.org/10.1068/p251009},
}

@article{doi:10.1167/12.1.12,
author = {Sun, Peng and Schofield, Andrew J.},
title = {Two operational modes in the perception of shape from shading revealed by the effects of edge information in slant settings},
journal = {Journal of Vision},
volume = {12},
number = {1},
pages = {12},
year = {2012},
%doi = {10.1167/12.1.12},

}

@inproceedings{zucker2019borders,
  title={From Borders to Bumps: Circular Flows Are Invariant Across Materials},
  author={Zucker, Steven and Kunsberg, Benjamin},
  booktitle={PERCEPTION},
  volume={48},
  pages={69--69},
  year={2019},
 % organization={SAGE PUBLICATIONS LTD 1 OLIVERS YARD, 55 CITY ROAD, LONDON EC1Y 1SP, ENGLAND}
}

@article{ramachandran1988perception,
  title={Perception of shape from shading},
  author={Ramachandran, Vilayanur S},
  journal={Nature},
  volume={331},
  number={6152},
  pages={163--166},
  year={1988},
  publisher={Nature Publishing Group}
}

@article{van2009filling,
  title={Filling-in afterimage colors between the lines},
  author={van Lier, Rob and Vergeer, Mark and Anstis, Stuart},
  journal={Current Biology},
  volume={19},
  number={8},
  pages={R323--R324},
  year={2009},
  publisher={Elsevier}
}

@book{pessoa2003filling,
  title={Filling-in: From perceptual completion to cortical reorganization},
  author={Pessoa, Luiz and De Weerd, Peter},
  year={2003},
  publisher={Oxford University Press}
}

@article{bressan1997neon,
  title={Neon color spreading: a review},
  author={Bressan, Paola and Mingolla, Ennio and Spillmann, Lothar and Watanabe, Takeo},
  journal={Perception},
  volume={26},
  number={11},
  pages={1353--1366},
  year={1997},
  publisher={SAGE Publications Sage UK: London, England}
}

@article{li2000perception,
  title={Perception of three-dimensional shape from texture is based on patterns of oriented energy},
  author={Li, Andrea and Zaidi, Qasim},
  journal={Vision research},
  volume={40},
  number={2},
  pages={217--242},
  year={2000},
  publisher={Elsevier}
}

@article{rao1991computing,
  title={Computing oriented texture fields},
  author={Rao, A Ravishankar and Schunck, Brian G},
  journal={CVGIP: Graphical Models and Image Processing},
  volume={53},
  number={2},
  pages={157--185},
  year={1991},
  publisher={Elsevier}
}

@article{bigun1991multidimensional,
  title={Multidimensional orientation estimation with applications to texture analysis and optical flow},
  author={Big{\"u}n, Josef and Granlund, Goesta H. and Wiklund, Johan},
  journal={IEEE Transactions on Pattern Analysis \& Machine Intelligence},
  number={8},
  pages={775--790},
  year={1991},
  publisher={IEEE}
}

@article{fleming2011estimation,
  title={Estimation of 3D shape from image orientations},
  author={Fleming, Roland W and Holtmann-Rice, Daniel and B{\"u}lthoff, Heinrich H},
  journal={Proceedings of the National Academy of Sciences},
  volume={108},
  number={51},
  pages={20438--20443},
  year={2011},
  publisher={National Acad Sciences}
}

@article{marlow2019photogeometric,
  title={Photogeometric cues to perceived surface shading},
  author={Marlow, Phillip J and Mooney, Scott WJ and Anderson, Barton L},
  journal={Current Biology},
  volume={29},
  number={2},
  pages={306--311},
  year={2019},
  publisher={Elsevier}
}

@article{Mooney14,
title = "Specular Image Structure Modulates the Perception of Three-Dimensional Shape ",
journal = "Current Biology ",
volume = "24",
number = "22",
pages = "2737 - 2742",
year = "2014",
author = "Scott W.J. Mooney and Barton L. Anderson",
}

@article{MAMASSIAN19962351,
title = "Illumination, Shading and the Perception of Local Orientation ",
journal = "Vision Research ",
volume = "36",
number = "15",
pages = "2351 - 2367",
year = "1996",
author = "Pascal Mamassian and Daniel Kersten"
}

@Article{Mingolla1986,
author="Mingolla, E.
and Todd, J. T.",
title="Perception of solid shape from shading",
journal="Biological Cybernetics",
year="1986",
volume="53",
number="3",
pages="137--151",
%url="http://dx.doi.org/10.1007/BF00342882"
}

@misc{marr1982vision,
  title={Vision: A computational investigation into the human representation and processing of visual information},
  author={Marr, David},
  year={1982},
  publisher={San Francisco: WH Freeman}
}

@article{koenderink1984does,
  title={What does the occluding contour tell us about solid shape?},
  author={Koenderink, Jan J},
  journal={Perception},
  volume={13},
  number={3},
  pages={321--330},
  year={1984},
  publisher={SAGE Publications Sage UK: London, England}
}

@article{cholewiak2014predicting,
  title={Predicting 3D shape perception from shading and texture flows},
  author={Cholewiak, Steven A and Kunsberg, Benjamin and Zucker, Steven and Fleming, Roland W},
  journal={Journal of Vision},
  volume={14},
  number={10},
  pages={1113--1113},
  year={2014},
  publisher={The Association for Research in Vision and Ophthalmology}
}

@article{holtmann2012superposition,
  title={Superposition of Glass Patterns: finding the flow through local measurements},
  author={Holtmann-Rice, Daniel and Ben-Shahar, Ohad and Zucker, Steven},
  journal={Journal of Vision},
  volume={12},
  number={9},
  pages={1312--1312},
  year={2012},
  publisher={The Association for Research in Vision and Ophthalmology}
}

@article{ben2004geometrical,
  title={Geometrical computations explain projection patterns of long-range horizontal connections in visual cortex},
  author={Ben-Shahar, Ohad and Zucker, Steven},
  journal={Neural computation},
  volume={16},
  number={3},
  pages={445--476},
  year={2004},
  publisher={MIT Press}
}

@inproceedings{huggins2001finding,
  title={Finding folds: On the appearance and identification of occlusion},
  author={Huggins, Patrick S and Chen, Hansen F and Belhumeur, Peter N and Zucker, Steven W},
  booktitle={Proceedings of the 2001 IEEE Computer Society Conference on Computer Vision and Pattern Recognition. CVPR 2001},
  volume={2},
  pages={II--II},
  year={2001},
  organization={IEEE}
}

@inproceedings{huggins2001folds,
  title={Folds and cuts: how shading flows into edges},
  author={Huggins, Patrick S and Zucker, Steven W},
  booktitle={Computer Vision, 2001. ICCV 2001. Proceedings. Eighth IEEE International Conference on},
  volume={2},
  pages={153--158},
  year={2001},
  organization={IEEE}
}

@article{palmer2008extremal,
  title={Extremal edge: A powerful cue to depth perception and figure-ground organization},
  author={Palmer, Stephen E and Ghose, Tandra},
  journal={Psychological science},
  volume={19},
  number={1},
  pages={77--83},
  year={2008},
  publisher={SAGE Publications Sage CA: Los Angeles, CA}
}

@article{Reininghaus11,
author="Reininghaus, Jan
and Hotz, Ingrid",
title="Combinatorial 2D Vector Field Topology Extraction and Simplification",
journal = "IEEE Trans Vis Comput Graph.",
year="2011",
publisher="Springer Berlin Heidelberg",
pages="1433--1443",
}

@article {Kunsberg2018Focus,
	author = {Kunsberg, Benjamin and Holtmann-Rice, Daniel and Alexander, Emma and Cholewiak, Steven and Fleming, Roland and Zucker, Steven W.},
	title = {Colour, contours, shading and shape: flow interactions reveal anchor neighbourhoods},
	volume = {8},
	number = {4},
	year = {2018},
	doi = {10.1098/rsfs.2018.0019},
	publisher = {Royal Society},
%	abstract = {Two dilemmas arise in inferring shape information from shading. First, depending on the rendering physics, images can change significantly with (even) small changes in lighting or viewpoint, while the percept frequently does not. Second, brightness variations can be induced by material effects{\textemdash}such as pigmentation{\textemdash}as well as by shading effects. Improperly interpreted, material effects would confound shading effects. We show how these dilemmas are coupled by reviewing recent developments in shape inference together with a role for colour in separating material from shading effects. Aspects of both are represented in a common geometric (flow) framework, and novel displays of hue/shape interaction demonstrate a global effect with interactions limited to localized regions. Not all parts of an image are perceptually equal; shape percepts appear to be constructed from image anchor regions.},
%	issn = {2042-8898},
%	URL = {http://rsfs.royalsocietypublishing.org/content/8/4/20180019},
%	eprint = {http://rsfs.royalsocietypublishing.org/content/8/4/20180019%.full.pdf},
	journal = {Interface Focus}
}

@article{zucker-sff,
	author = "Breton, P. and Zucker, S.W.",
	title = "Shadows and Shading Flow Fields",
	journal = "IEEE Conf. on Computer Vision and Pattern Recognition",
	address = "San Francisco, CA",
	month = "June",
	year = "1996",
	pages = "782 - 789"
}

@article{freeman94,
	author = "W. Freeman",
	title = "The generic viewpoint assumption in a framework for visual perception",
	journal = "Nature",
	year = "1994",
	vol="368",
	pages = "542 - 545"
}

@article{Freeman96,
author = {Freeman, William T},
title = {{Exploiting the generic viewpoint assumption}},
journal = {International Journal Of Computer Vision},
year = {1996},
volume = {20},
number = {3},
pages = {243--261}
}

@article{Giorgis15,
	author = "Allemand-Giorgis, L. and Bonneau, G-P. and Hahmann, S.",
	title = "{P}iecewise {P}olynomial {R}econstruction of {F}unctions from {S}implified {M}orse-{S}male complex",
	journal = "IEEE Visualization Conference",
	year = "2014",
}

@article{Weinkauf10,
	author = "Weinkauf, T. and Gingold, Y. and Sorkine, O.",
	title = "{T}opology-based {S}moothing of 2{D} {S}calar {F}ields with {C1}-{C}ontinuity",
	journal = "IEEE Symposium on Visualization",
	year = "2010",
}

@article {Weinkauf09,
author = {Weinkauf, T. and Gunther, D.},
title = {Separatrix Persistence: Extraction of Salient Edges on Surfaces Using Topological Methods},
journal = {Computer Graphics Forum},
volume = {28},
number = {5},
publisher = {Blackwell Publishing Ltd},
%issn = {1467-8659},
%url = {http://dx.doi.org/10.1111/j.1467-8659.2009.01528.x},
%doi = {10.1111/j.1467-8659.2009.01528.x},
pages = {1519--1528},
keywords = {G.2.3 [Mathematics of Computing]: Discrete Mathematics, Applications I.3.0 [Computer Graphics]: General},
year = {2009},
}

@article{Sahner08,
	author = "Sahner, J. and Weber, B. and Prohaska, S. and Lamecker, H.",
	title = "{E}xtraction of {F}eature {L}ines on {S}urface {M}eshes based on {D}iscrete {M}orse {T}heory",
	journal = "IEEE Symposium on Visualization",
	volume = "27",
	number = "3",
	year = "2008",
}

@book{oneill,
        author = "Barrett O'Neill",
        title = "Elementary Differential Geometry, Revised 2nd Edition",
        publisher = "Elsevier",
        address = "Burlington, Massachusetts",
        year = "2006",
}

@book{horn89,
        author = "B.K.P. Horn and M.J. Brooks",
        title = "Shape from Shading",
        publisher = "The MIT Press",
        address = "Cambridge, Massachusetts",
        year = "1989",
}

@article{dakin2002summation,
  title={Summation of concentric orientation structure: seeing the Glass or the window?},
  author={Dakin, SC and Bex, PJ},
  journal={Vision Research},
  volume={42},
  number={16},
  pages={2013--2020},
  year={2002},
  publisher={Elsevier}
}

@article{gallant2000human,
  title={A human extrastriate area functionally homologous to macaque V4},
  author={Gallant, Jack L and Shoup, Rachel E and Mazer, James A},
  journal={Neuron},
  volume={27},
  number={2},
  pages={227--235},
  year={2000},
  publisher={Elsevier}
}

@article{fleming2004specular,
  title={Specular reflections and the perception of shape},
  author={Fleming, Roland W and Torralba, Antonio and Adelson, Edward H},
  journal={Journal of vision},
  volume={4},
  number={9},
  pages={10--10},
  year={2004},
  publisher={The Association for Research in Vision and Ophthalmology}
}

@article{dumoulin2007cortical,
  title={Cortical specialization for concentric shape processing},
  author={Dumoulin, Serge O and Hess, Robert F},
  journal={Vision research},
  volume={47},
  number={12},
  pages={1608--1613},
  year={2007},
  publisher={Elsevier}
}

@article{achtman2003sensitivity,
  title={Sensitivity for global shape detection},
  author={Achtman, Rebecca L and Hess, Robert F and Wang, Yi-Zhong},
  journal={Journal of Vision},
  volume={3},
  number={10},
  pages={4--4},
  year={2003},
  publisher={The Association for Research in Vision and Ophthalmology}
}

@article{wilson1998detection,
  title={Detection of global structure in Glass patterns: implications for form vision},
  author={Wilson, Hugh R and Wilkinson, Frances},
  journal={Vision research},
  volume={38},
  number={19},
  pages={2933--2947},
  year={1998},
  publisher={Elsevier}
}

@article{kovacs1993closed,
  title={A closed curve is much more than an incomplete one: Effect of closure in figure-ground segmentation},
  author={Kovacs, Ilona and Julesz, Bela},
  journal={Proceedings of the National Academy of Sciences},
  volume={90},
  number={16},
  pages={7495--7497},
  year={1993},
  publisher={National Acad Sciences}
}

@article{elder1993effect,
  title={The effect of contour closure on the rapid discrimination of two-dimensional shapes},
  author={Elder, James and Zucker, Steven},
  journal={Vision research},
  volume={33},
  number={7},
  pages={981--991},
  year={1993},
  publisher={Pergamon}
}

@book{koenderink90Koen,
        author = "Jan J Koenderink",
        title = "Solid {S}hape",
        publisher = "The MIT Press",
        address = "Cambridge, Massachusetts",
        year = "1990",
}

@article{Connor08,
        author = "Yukako Yamane and Eric T Carlson and Katherine C Bowman and Zhihong Wang and Charles E Connor",
        title = "A {N}eural {C}ode for {T}hree-{D}imensional {O}bject {S}hape in {M}acaque {I}nferotemporal {C}ortex",
        journal = "Nature Neuroscience: Published Online",
        year = "2008",
}

@phdthesis{Gyulassy08,
  title={Combinatorial construction of Morse-Smale complexes for data analysis and visualization},
  author={Gyulassy, Attila Gabor},
  year={2008},
  school={University of California, Davis}
}

@article{ToddVSS17,
author = "Nartker, M. and Todd, J. and Petrov, A.",
title = "Distortions of apparent 3D shape from shading caused by changes in the direction of illumination",
journal = "Journal of Vision",
year =  "2017",
volume = "17",
number = "324"
}

@article{koenderink15Koe,
author = {Jan Koenderink and Andrea van Doorn and Johan Wagemans},
title = {Part and Whole in Pictorial Relief},
journal = {i-Perception},
volume = {6},
number = {6},
year = {2015},
URL = { 
        http://dx.doi.org/10.1177/2041669515615713}
}

@article{Judd07,
  author    = {Tilke Judd and
               Fredo Durand and
               Edward H. Adelson},
  title     = {Apparent ridges for line drawing},
  journal   = {ACM Trans. Graph.},
  volume    = {26},
  number    = {3},
  year      = {2007},
  pages     = {19},
}

@article{holtmann2018tensors,
  title={Tensors, Differential Geometry and Statistical Shading Analysis},
  author={Holtmann-Rice, Daniel N and Kunsberg, Benjamin S and Zucker, Steven W},
  journal={Journal of Mathematical Imaging and Vision},
  volume={60},
  number={6},
  pages={968--992},
  year={2018},
  publisher={Springer}
}

@article{Kunsberg18,
author = {Kunsberg, Benjamin and Zucker, Steven W},
title = {Critical Contours: An Invariant Linking Image Flow with Salient Surface Organization},
journal = {SIAM Journal on Imaging Sciences},
year = {2018},
}

@article{DeCarlo03,
  author = "Doug DeCarlo and Adam Finkelstein and Szymon Rusinkiewicz and Anthony Santella",
  title = "Suggestive Contours for Conveying Shape",
  journal = "ACM Trans. Graph.",
  year = "2003",
  volume = "22",
  number = "3",
  pages = "848--855"
}

@article{Lawlor200918,
title = "Boundaries, shading, and border ownership: A cusp at their interaction ",
journal = "Journal of Physiology-Paris ",
volume = "103",
number = "12",
pages = "18 - 36",
year = "2009",
author = "Matthew Lawlor and Daniel Holtmann-Rice and Patrick Huggins and Ohad Ben-Shahar and Steven W. Zucker"
}

@article{Biasotti:2008:DSG:1391729.1391731,
 author = {Biasotti, S. and De Floriani, L. and Falcidieno, B. and Frosini, P. and Giorgi, D. and Landi, C. and Papaleo, L. and Spagnuolo, M.},
 title = {Describing Shapes by Geometrical-topological Properties of Real Functions},
 journal = {ACM Comput. Surv.},
 issue_date = {October 2008},
 volume = {40},
 number = {4},
 month = oct,
 year = {2008},
 issn = {0360-0300},
 pages = {12:1--12:87},
 articleno = {12},
 numpages = {87},
 url = {http://doi.acm.org/10.1145/1391729.1391731},
 acmid = {1391731},
 publisher = {ACM},
 address = {New York, NY, USA},
}

@article{carlsson2009topology,
  title={Topology and data},
  author={Carlsson, Gunnar},
  journal={Bulletin of the American Mathematical Society},
  volume={46},
  number={2},
  pages={255--308},
  year={2009}
}

@article{koenderink:1980bm,
author = {Koenderink, Jan J and van Doorn, Andrea J},
title = {{Photometric invariants related to solid shape}},
journal = {Journal of Modern Optics},
year = {1980},
volume = {27},
number = {7},
pages = {981--996}
}

@book{milnor2016morse,
  title={Morse Theory},
  author={Milnor, John},
  year={1963},
  publisher={Princeton University Press}
}

@article{zhou2000coding,
  title={Coding of border ownership in monkey visual cortex},
  author={Zhou, Hong and Friedman, Howard S and Von Der Heydt, R{\"u}diger},
  journal={Journal of Neuroscience},
  volume={20},
  number={17},
  pages={6594--6611},
  year={2000},
  publisher={Soc Neuroscience}
}

@book{matsumoto2002introduction,
  title={An introduction to Morse theory},
  author={Matsumoto, Yukio},
  volume={208},
  year={2002},
  publisher={American Mathematical Soc.}
}
\end{document}


\title{Supplementary Material}
\author{Benjamin Kunsberg and Steven W. Zucker\\
\emph{Applied Mathematics, Brown University}}
\maketitle

\section{Suggestive Contours and Apparent Ridges}
Suggestive contours and apparent ridges are similar ideas, yet share two fundamental differences with our approach.  First, they are computed from the surface and there is currently no understood relationship between those surface contours and image patterns.  Second, they are defined with a pointwise condition and so do not necessarily delineate intuitive regions of the surface.  For example, there is no theory relating the suggestive contours to global surface shape and so there is no way to interpolate back the surface from suggestive contours.  Rather, we use extremal image curves as part of a larger global complex, and we show that they can be related to global surface shape.

\section{Example of Reconstruction from MS}


\begin{figure}[ht]
\begin{center}
\includegraphics[trim={1cm 2cm 7cm 2.3cm},clip, width=0.8 \linewidth]{images/complete_bunny.pdf}
\caption{These images illustrate how function values on the curves of the Morse Smale (MS) complex can be interpolated to approximate a function.  Upper left: A shaded image of the Stanford Bunny.  Center column: The MS complex at two different levels of persistence.  Right column:  Harmonic interpolation using the shading values on the MS complex and the boundary.  We see that qualitatively the images are similar to the original.  We argue similarly that the MS complex of the slant function is a good qualitative representation of the surface slant function. \label{fig:bunny}}
\end{center}
\end{figure}

\begin{figure}[h]
\begin{center}
\includegraphics[trim={2cm 0 0cm 0},clip,width=0.8 \linewidth]{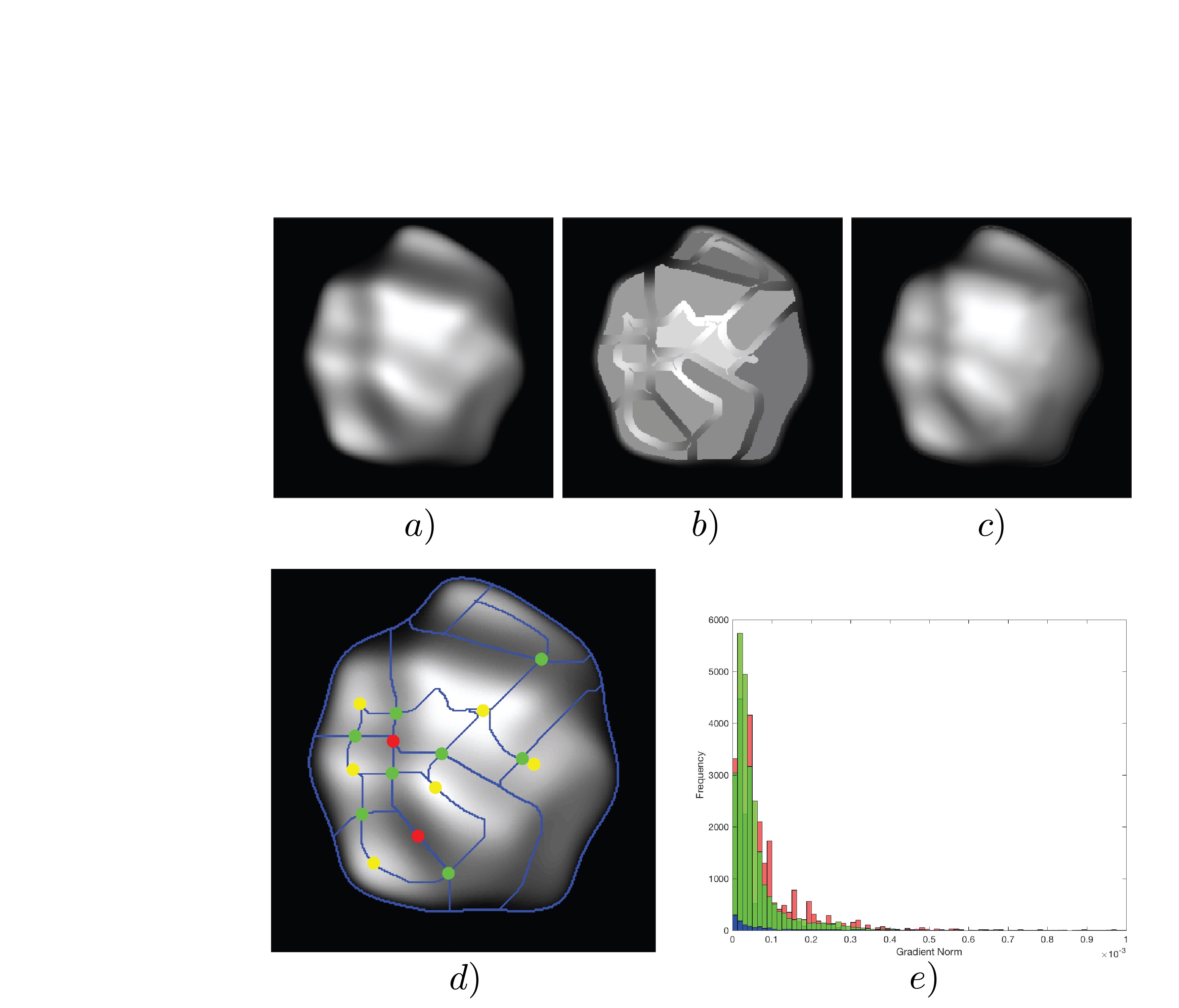}
\caption{a) An image of a `blob' with a vivid shape percept.  b) A `flattened' image computed from a) where the intensities on the extremal image lines are kept equivalent and, for each region, we use the mean intensity of the region.  c) A blurring of b).  d) The extremal lines and critical points of image a). e) A histogram comparing gradient norms for images c) in green, b) in blue, and a) in red. This shows that c) and a) are not  the same images although they may be perceived similarly; c) has smaller gradients than a) in general. \label{fig:inpainting}}
\end{center}
\end{figure}

We expand upon the example in Figure ?? of the Stanford Bunny.  We wish to illustrate the fact that the extremal lines of an image can, in fact, represent the majority of the percievable structure in the image. In Figure \ref{fig:inpainting}, we illustrate the process of `inpainting' an image from its extremal lines.

We start with an image of a `blob' in Fig \ref{fig:inpainting}(a). We compute the image extremal curves (the MS complex), shown in Fig \ref{fig:inpainting}(d).  These extremal curves run through the critical points of the image.  Red points correspond to minima, yellow points correspond to maxima, and green points correspond to saddles.  We now take the pixel values on the MS complex and set the vast majority of the image to constant values shown in Fig \ref{fig:inpainting}(b).  We then apply blurring via the heat kernel to get Fig \ref{fig:inpainting}(c).  We see that the quantiative pixel values on the extremal lines can `sparsely' represent the original image with nearly imperceptibly differences.  We compare the histogram of gradients of the original image and the reconstruction in Fig \ref{fig:inpainting}(e).  Here, we see that (e) is a smoother image (lower gradient norms) than (a) and so the images are different even if the qualitative perceptions are similar.

Suppose that the visual system extracted the extermal curves in (b) from the original image (a) as we are advocating.  This would not only be a sparser representation; it would also have minimal dependence on the illuminance and rendering function as shown in \cite{Kunsberg18}.

\section{Generic Argument against the Twisted Surface}

We remarked in Section 3.1 that there was a generic argument that the normal field along a slant extremal curve must uniformly `point to one side'.  We now give this argument, following Freeman's approach in \cite{freeman94}.  The gist of the argument is one can define a probability distribution on solutions to a constraint based on the stability of `nuisance parameters'.  Here, we take the nuisance parameter to be the direction of illumination.  The most generic solutions then will be the solutions that are most stable with respect to changes in the illumination.

Define, using the notation in \cite{freeman94},
\begin{align}
\bm{\beta} & = \D \n \\
\bm{Y} & =  \D I 
\end{align}

Now, consider the following rendering function $\mathbf{g}$:

\begin{align}
\mathbf{g}(\l, \bm{\beta} ) & = \bm{\hat{Y}} \\
& = \l \tr \bm{\beta}
\end{align}

Following \cite{freeman94, Freeman96}, we call $\l$ the \emph{generic variable}, $\bm{\beta}$ the \emph{scene parameters}, and $\bm{Y}$ the \emph{observations}.  $\l$ is an unknown vector in $\mathbb{R}^3$.  Following \cite{freeman94, Freeman96}, we assume a Gaussian noise model on the observations $\bm{Y}$: 
\begin{align}
\bm{Y} = \hat{\bm{Y}} + \bm{T}
\end{align}
 
\noindent
where $\hat{\bm{Y}}$ is the ideal rendered observation and $T \sim N(0, \bm{\Sigma})$ with $\bm{\Sigma} = \text{diag}(\sigma^2)$ for some $\sigma \in \mathbb{R}^+$.  For the noise model, we have

\begin{align}
P(\bm{Y} | \bm{\beta}, \l) & =  \frac{1}{(\sqrt{2 \pi \sigma^2})^m} e^{- \frac{ || \bm{Y} -\mathbf{g}(\l, \bm{\beta}) ||^2}{2 \sigma^2}}
\end{align}

\noindent
This is the probability of the surface $\D \n$ given a noisy observation of the image gradient $\D I$.  Applying Bayes' theorem and integrating over the nuisance variable $\l$ yields the posterior distribution:

\begin{align}
P(\bm{\beta} | \bm{Y}) & = k \exp \left( \frac{ - ||\bm{Y} - \mathbf{g}(\l_0 , \bm{\beta}) ||^2}{2 \sigma^2} \right) \nonumber \\
& \cdot [ P_{\bm{\beta}} (\bm{\beta}) P_{\l} (\l_0)] \frac{1}{\sqrt{\det (\bm{A})}} \label{eqn:posterior}\\
& = k \hspace{2mm} \text{(fidelity)} \nonumber \\
& \cdot \hspace{2mm} \text{(prior probability)} \hspace{2mm} \text{(genericity)} \nonumber
\end{align}
\noindent
where $P_{\bm{\beta}} (\bm{\beta}), P_{\l} (\l_0)$ are prior distributions on the surface and light source parameters, $\l_0$ is the light source that can best account for the observations given a chosen $\bm{\beta}$ and $\bm{A}$ is a matrix with the following elements:

\begin{equation}
A_{ij} = \mathbf{g}'_i \cdot \mathbf{g}'_j - (\bm{Y} - \mathbf{g}(\l_0, \bm{\beta})) \cdot \mathbf{g}''_{ij}
\end{equation}
with 
\begin{align}
\mathbf{g}'_i & = \frac{\partial \mathbf{g} (\l, \bm{\beta})}{\partial l_i} \Big|_{\l = \l_0} \\
\mathbf{g}''_{ij} & = \frac{\partial^2 \mathbf{g} (\l, \bm{\beta})}{\partial l_i \partial l_j} \Big|_{\l = \l_0}
\end{align}

For more details of the previous Bayesian analysis, consult \cite{Freeman96}. We now seek the solutions that maximize the posterior probability $P(\bm{\beta} | \bm{Y})$.  From (\ref{eqn:posterior}), we maximize by choosing $\bm{\beta}$ and $\l_0$ so that $||\bm{Y} - \mathbf{g}(\l_0 , \bm{\beta}) || = 0$ while at the same time setting $\det(\bm{A}) = 0$.  

This means we can maximize the probability of the surface property $\D \n$  given the image gradient $\D I$ if we set $\det(\bm{A}) = 0$.  This is equivalent to ensuring that $\D \n$ is low rank and so the surface has zero (or minimal) Gaussian curvature in the given patch.  As shown in Fig ?? in the main paper, we see that the Gaussian curvature is very large for the twisted surface whereas the Gaussian image is 0 for the non-twisted surface.


\section{Depth Functions for Figure \ref{fig:spherical_ridge}}

We show the height functions for the different labeling cases in Figure ??a and Figure ??d in the main paper.
\begin{figure}[h]
\begin{center}
\includegraphics[width= 0.4 \linewidth]{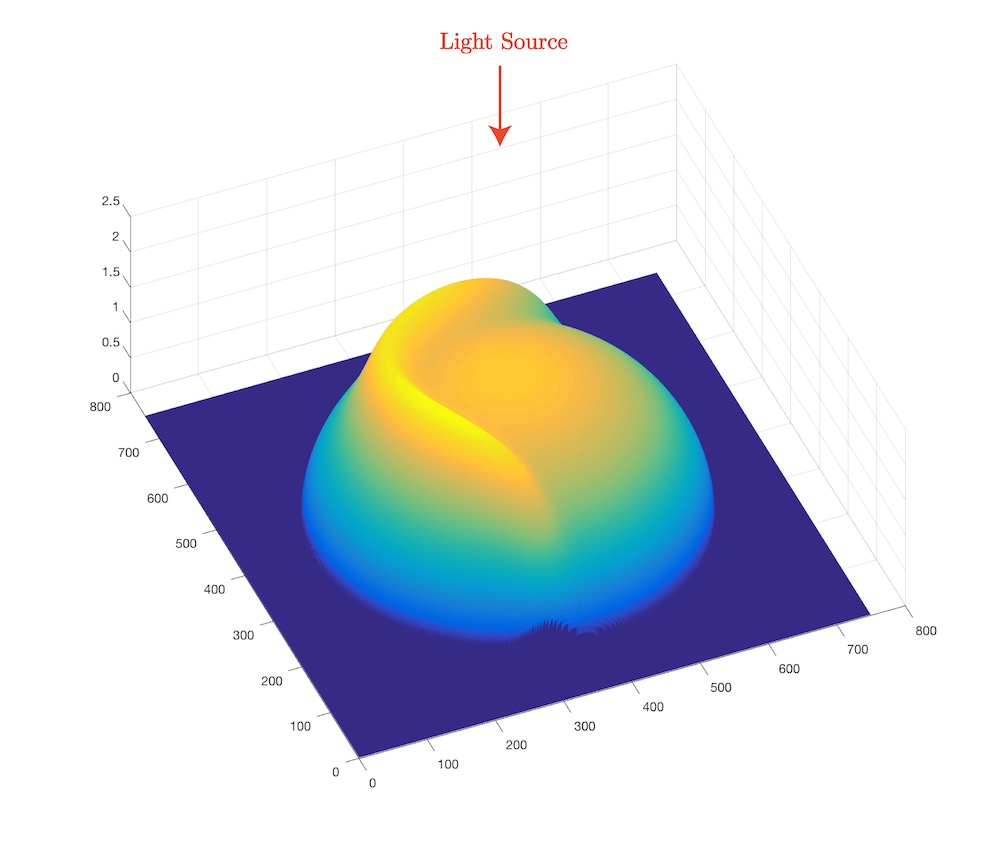}
\includegraphics[width= 0.4 \linewidth]{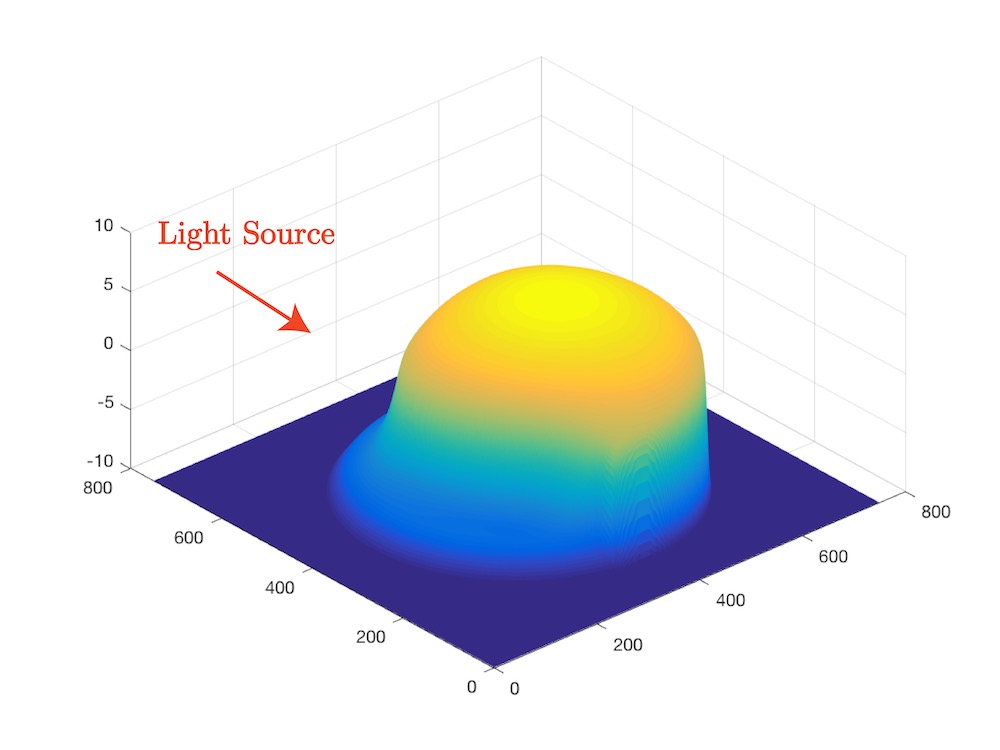}
\caption{Left: Surface corresponding to the topology described in labeling (a) in Fig \ref{fig:spherical_ridge}. Right: Surface corresponding to the topology described in labeling (d) in Fig \ref{fig:spherical_ridge}.  Note that diagrams b) and c) are the negative height functions of a) and d).  Left is more generic than Right.  You can think of Right as starting with Left and pulling the right side of the surface up until it is above the ridge.}
\end{center}
\end{figure}

\section{The Dual Ridge example}

We can extrapolate the ideas illustrated in Fig ?? to multiple ridges.  Here is an example of the extremal curves for a `two-ridge spherical coin'.  The possible labelings can clearly be deduced in the same manner as Fig ??.   

\begin{figure}[h]
\begin{center}
\includegraphics[width=1 \linewidth]{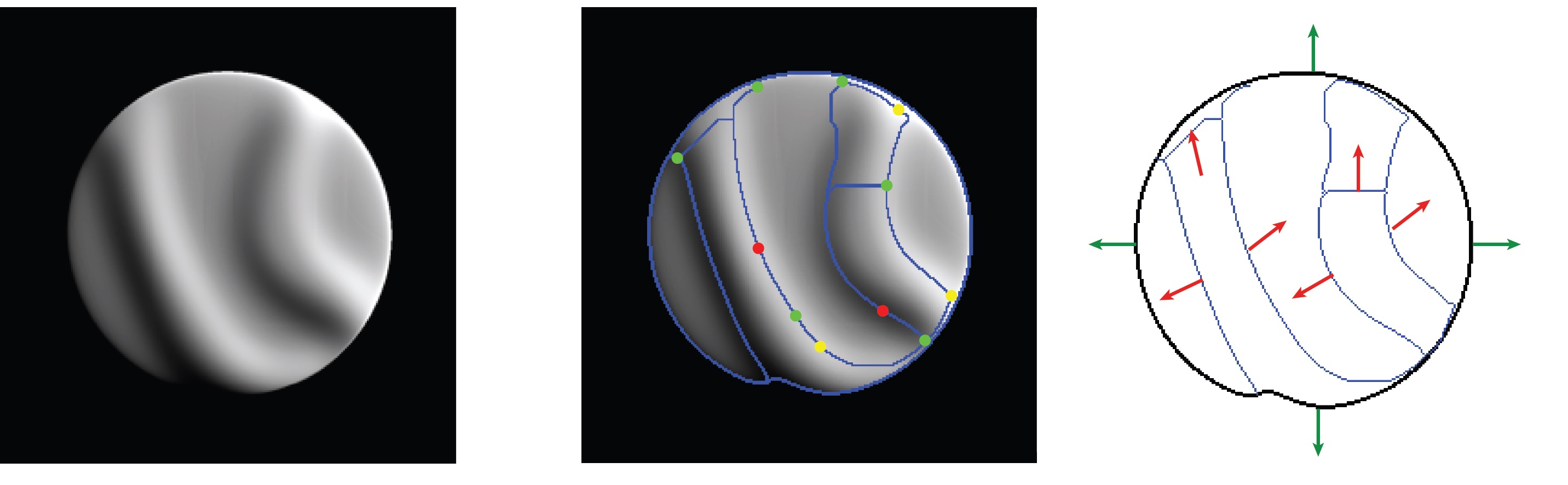}
\caption{Left: An image of the `two-ridge spherical coin.'  Center: The computed extremal image curves.  Right: The author's perceived labeling.}
\end{center}
\end{figure}

\newpage

\printbibliography